\documentclass[12pt,a4paper,twoside]{article}
\usepackage[margin=1in,headheight=15pt]{geometry}

\usepackage{natbib}

\usepackage[breaklinks=true]{hyperref}
\usepackage{breakcites}

\usepackage{graphicx}
\usepackage{amssymb,amsmath,amsfonts}
\usepackage{bm}
\usepackage{graphicx} 
\usepackage{bbm}
\usepackage{placeins}
\usepackage{amsfonts}
\usepackage{booktabs} 
\usepackage{array} 

\usepackage{algorithm}
\usepackage{setspace}

\newcommand{\smallbullet}{\vcenter{\hbox{\tiny$\bullet$}}}
\usepackage{xcolor,colortbl,caption}
\usepackage[FIGTOPCAP]{subfigure}

\usepackage{hyperref}

\newcommand{\Enc}{\mbox{\tt Enc}}
\newcommand{\Dec}{\mbox{\tt Dec}}
\newcommand{\argmax}{\operatornamewithlimits{arg\,max}}

\usepackage{fancyvrb}
\DefineVerbatimEnvironment{Highlighting}{Verbatim}{commandchars=\\\{\}}
\usepackage{framed}
\definecolor{shadecolor}{RGB}{248,248,248}
\newenvironment{Shaded}{\begin{snugshade}}{\end{snugshade}}
\newcommand{\KeywordTok}[1]{\textcolor[rgb]{0.13,0.29,0.53}{\textbf{{#1}}}}
\newcommand{\DataTypeTok}[1]{\textcolor[rgb]{0.13,0.29,0.53}{{#1}}}
\newcommand{\DecValTok}[1]{\textcolor[rgb]{0.00,0.00,0.81}{{#1}}}

\newcommand{\StringTok}[1]{\textcolor[rgb]{0.31,0.60,0.02}{{#1}}}

\newcommand{\NormalTok}[1]{{#1}}

\usepackage{caption}

    \def\independenT#1#2{\mathrel{\setbox0\hbox{$#1#2$}%
    \copy0\kern-\wd0\mkern4mu\box0}}

\usepackage{listings}
\usepackage{fancyhdr}
\title{Encrypted statistical machine learning: new privacy preserving methods}
\author{Louis~J.~M.~Aslett, Pedro~M.~Esperan\c{c}a and Chris~C.~Holmes  \\ {\small Department of Statistics, University of Oxford}}
\date{}

\begin{document}

\maketitle

\begin{abstract}
We present two new statistical machine learning methods designed to learn on fully homomorphic encrypted (FHE) data. The introduction of FHE schemes following \citet{Gentry09} opens up the prospect of privacy preserving statistical machine learning analysis and modelling of encrypted data without compromising security constraints. We propose tailored algorithms for applying extremely random forests, involving a new cryptographic stochastic fraction estimator, and na\"{i}ve Bayes, involving a semi-parametric model for the class decision boundary, and show how they can be used to learn and predict from encrypted data. We demonstrate that these techniques perform competitively on a variety of  classification data sets and provide detailed information about the computational practicalities of these and other FHE methods.

\vspace{5mm}
\end{abstract}

\section{Introduction}

Privacy requirements around data can impede the uptake and application of statistical analysis and machine learning algorithms.  Traditional cryptographic methods enable safe long-term storage of information, but when analysis is to be performed the data must first be decrypted.  \citet{Rivest78} initially showed that it may be possible to design an encryption scheme that supports restricted mathematical computations without decrypting. However, it was not until \citet{Gentry09} that a scheme able to support theoretically arbitrary computation was proposed. Briefly here, these so-called \emph{homomorphic encryption} schemes allow for certain mathematical operators such as addition and multiplication to be performed directly on the cipher texts (encrypted data), yielding encrypted results which upon decryption render the same results as if the operations had been performed on the plain texts (original data). These schemes are reviewed in a companion report to this paper \citep*{Part1} in a manner which is accessible to statisticians and machine learners with accompanying high level open source software in R to allow users to explore the various issues\footnote{In this report we will assume that the reader is familiar with the basic concepts of fully homomorphic encryption and some of the practical computational constraints, as overviewed in \citet*{Part1} and \citet{Gentry10}.}.

Privacy constraints enter into many areas of modern data analysis from biobanks and medical data to the impending wave of `wearable devices' such as smart watches, generating large amounts of personal biomedical data \citep{anderlik2001privacy,Kaufman09, angrist2013genetic,brenner2013prepared, ginsburg2014medical}.  Moreover with the advent of cloud computing many data owners are looking to outsource storage and computing, but particularly with non-centralised services there may be concerns with security issues during data analysis \citep{liu2011data}.  Indeed, encryption may even be desirable on internal network connected systems as providing an additional layer of security.

Although homomorphic encryption in theory promises arbitrary computation, the practical constraints mean that this is presently out of reach for many algorithms \citep*[][]{Part1}.  This motivates the interest in tailored machine learning methods which can be practically applied. This paper contributes two such methods with FHE approximations to extremely random forests and na\"{i}ve Bayes developed, such that both learning and prediction can be performed encrypted, something which is not possible with the original versions of either technique.

We are not the first to explore secure machine learning approaches to encryption. \citet{GLN12} implemented two binary classification algorithms for homomorphically encrypted data: Linear Means and Fisher's Linear Discriminant.  They make scaling adjustments which preserve the results, but leave the fundamental methodology unchanged.
\cite{BostETAL14} developed a two party computation framework and used a mix of different partly and fully homomorphic encryption schemes which allows them to use machine learning techniques based on hyperplane decisions, na\"{i}ve Bayes and binary decision trees --- again the fundamental methodologies are unchanged, but here substantial communication between two (`honest but curious') parties is required.

These are two existing approaches to working within the constraints imposed by homomorphic encryption: either by the use of existing methods amenable to homomorphic computation; or by invoking multi-party methods.  Here, we consider tailored approximations to two statistical machine learning models which make them amenable to homomorphic encryption, so that all stages of fitting and prediction can be computed encrypted.  Thus, herein we contribute two machine learning algorithms tailored to the framework of fully homomorphic encryption and provide an R package implementing them \citep{ESpkg}.  These techniques do not require multi party communication.

Aside from classification techniques, other privacy preserving statistical methods have been proposed in the literature such as \mbox{small-$P$} linear regression ($P \leq 5$; \citealp{WH12}) and predictive machine learning using pre-trained models (e.g., logistic regression; \citealp{BLN14})

In Section 2 a brief recap of homomorphic encryption and consequences for data representation is presented, with the unfamiliar reader directed to \citet*{Part1} for a fuller review.  Section 3 contains a novel implementation of extremely random forests  \citep{Geurts06,Cutler01} including a stochastic approximation to tree voting. In Section 4 a novel semi-parametric na\"{i}ve Bayes algorithm is developed that utilises logistic regression to define the decision boundaries. Section 5 details empirical results of classification performance on a variety of tasks taken from the UCI machine learning repository, as well as demonstrating the practicality with performance metrics from fitting a completely random forest using the Amazon EC2 cloud platform. Section 6 offers a discussion and conclusions.

\section{Homomorphic encryption and data representation}
\label{sec:HE}

We shall adopt a {\em{public key encryption scheme}} having public key $k_p$ and secret key $k_s$ and equipped with algorithms $\Enc(k_p, \cdot)$ and $\Dec(k_s, \cdot)$ which encrypt and decrypt messages respectively.  Encryption maps a message $m \in M$ from message space to an element of cipher text space $c \in C$.  A scheme is then said to be homomorphic for some operations $\circ \in \mathcal{F}_M$ acting in message space (such as addition or multiplication) if there are corresponding operations $\diamond \in \mathcal{F}_C$ acting in cipher text space satisfying the property:
\[ \Dec(k_s, \Enc(k_p, m_1) \diamond \Enc(k_p, m_2)) = m_1 \circ m_2 \quad \forall\,m_1, m_2 \in M \]
A scheme is \emph{fully} homomorphic if it is homomorphic for both addition and multiplication. We shall consider herein the particular homomorphic encryption scheme of \citet{Fan12}, a high performance and easy to use implementation of which is available in R \citep{HEpkg}, and assume that the reader is familiar with the basic principles of this approach \citep*{Part1}.  

\subsection{Practical limitations}
\label{sec:limitations}

Although FHE schemes exist, it is worth briefly recalling the practical constraints in implementing arbitrary algorithms, as they impact and motivate the tailored developments presented in this paper. Some of the current practical implementation issues include:

\noindent\textbf{Message space:} Real value encryption lies outside of existing FHE schemes so that measurements must typically be stored as integers. Given an integer measurement, $x$, the choice of the corresponding message space representation, $M$, will have consequences for computational cost and memory requirements.  For example, $x$, could be  directly represented in an integer message space $M \subset \mathbb{Z}$, or in a binary message space $M=\{0,1\}^T$ involves writing down the value in base 2 ($x=\sum 2^i b_i$) and encrypting each bit (each $b_i$) separately.

The major consequence is that performing simple operations such as addition and multiplication under the binary representation involves manual binary arithmetic, which is much more expensive than the single operation involved when the natural integer representation is used.  For instance, adding two 32-bit values in a binary representation would involve over 256 fundamental operations by using standard full binary adder logic. Consequently, we do not consider FHE schemes where binary representation is the only option and instead require $M \subset \mathbb{Z}$ for the new techniques to be presented in Sections \ref{sec:RF} and \ref{sec:NaiveBayes}.

However, although $M \subset \mathbb{Z}$ is more efficient computationally, it still does not naturally accommodate the kinds of data commonly encountered in statistics and machine learning applications, so that even representing data requires careful consideration.

\noindent\textbf{Cipher text size:} existing FHE schemes result in substantial inflation in the size of the data.  For example, in \citet{Fan12} the cipher text space is $C = \mathbb{Z}_q[x] \times \mathbb{Z}_q[x]$, a cartesian product of high degree polynomial rings with coefficients belonging to a large integer ring.  Therefore when using the default parameter values in that paper the two polynomials are of degree $4,095$, with each of the $8,192$ coefficients being 128-bit integers.  This means that 1MB of message data can grow to approximately 16.4GB of encrypted data, representing a 1,600 fold increase in storage size.

\noindent\textbf{Computational speed:} due in part to the increased data size, but also due to the complex cipher text spaces, the cost of performing operations is high. For example, in \citet{Fan12} arithmetic for simple messages in $M$ is achieved by performing complex polynomial arithmetic in $C$.

To make this concrete, imagine adding the numbers 2 and 3 to produce 5.  Basic parameter choices for the \citet{Fan12} encryption scheme will mean that not only does this simple addition involve adding $4,095$ degree polynomials, but the 128-bit integer coefficients of those polynomials are too large to be natively represented or operated on by modern CPUs.

Indeed, the theoretical latency for integer addition on a modern CPU is 1 clock cycle, so that 2+3 executes in sub 1 nanosecond ($10^{-9}$s).  By contrast, the optimised C++ implementation of \citet{Fan12} in \citet{HEpkg} takes around 3 milliseconds ($10^{-3}$s) to perform the same computation encrypted.

\noindent\textbf{Division and comparison:} existing integer message space schemes cannot perform encrypted division and are unable to evaluate binary comparison operations such as $=, <$ and $>$. So that mathematical operations are currently restricted to addition and multiplication.

\noindent\textbf{Cryptographic noise:} the semantic security necessary in existing schemes involves injection of some noise into the cipher texts, which grows as operations are performed.  Typically the noise growth under multiplication can be significant so that after a certain depth of multiplications the cipher text must be `refreshed'.  This refresh step is usually computationally expensive, so that in practice the parameters of the encryption scheme are usually chosen a priori to ensure that all necessary operations for the algorithm to be applied can be performed without any refresh being required.

Thus, the restriction to integers, addition and multiplication, combined with a limit on noise growth emanating from multiplication operations, means that in reality the constraints of homomorphic encryption allow only moderate degree polynomials of integers to be computed encrypted.  Even so, the speed of evaluation will be relatively slow compared to the unencrypted counterparts, as demonstrated in our examples in Section 5.

\subsection{Data representation}
\label{sec:datarep}

One consequence of the above is that we need to transform data to make it amenable to FHE analysis. We show that certain transformations will also allow for limited forms of computation involving comparison operations such as $=, <$ and $>$. We consider two simple approaches below.

\subsubsection{Quantisation for real values}
\label{sec:EncodingReals}
Given that many current homomorphic schemes work in the space of integers \citep*{Part1}, it may be necessary to make approximations when manipulating real-valued variables. 
\cite{GLN12} proposed an approximation method where real values are first approximated by rationals (two integers factors) and then denominators cleared---by multiplying the entire dataset by a pre-specified integer---and rounding the results to the nearest integer.

One suggestion here is more straightforward: choose a desired level of accuracy, say $\phi$, which represents the number of decimal places to be retained; then multiply the data by $10^{\phi}$ and round to the nearest integer.
This avoids the need for rational approximations and the requirement for a double approximation caused by the denominator-clearing step. More precisely, for a given precision $\phi \in \mathbb{N} \cup \{0\}$, a real value $z \in \mathbb{R}$ is approximated by $\dot{z} = \lfloor 10^{\phi} \cdot z \rceil$, where $\lfloor \cdot \rceil$ denotes rounding to the nearest integer.
This transformation adequately represents real values in an integer space, in the sense that smooth relative distances are approximately maintained.

For data sets of finite precision (the typical case in real applications), no loss of precision is necessary if $\phi$ is selected to be equal to the accuracy (i.e., number of decimal places) of the most accurate value in the data set. Otherwise, in the cases where transformations are required (e.g., logarithms), precision is under the user's control.
Note that the parameter $\phi$ regulates the accuracy of the input (data), not that of the output (result). To the extent that the output accuracy depends on the input accuracy and also on the complexity of the algorithm, the choice of $\phi$ should take both these factors into consideration.

In particular, when evaluating homogeneous polynomial expressions, then no intermediate scaling is required since every term will have scaling $10^{d\phi}$, where $d$ is the degree of the homogeneous polynomial.  Where scaling is required, it will be known a priori based on the algorithm and is not data dependent.

\subsubsection{Quantisation to categorical or ordinal}
\label{sec:Quantisation}

The approach above encodes real values by an integer representation, but this increases substantially the number of multiplication operations involved.  Instead, by transforming continuous measurement values into categorical or ordinal ones via a quantisation procedure, it is possible to dispense with the need to track appropriate scaling in the algorithm. This simple solution has not, as far as we're aware, been taken in the applied cryptography literature to date.  Moreover, this quantisation procedure allows some computations involving comparison operations ($=, <, >$) to now be performed as detailed below.

Let $X$ be a design matrix, with elements $x_{ij}$ recording the $j$th predictor variable for the $i$th observation.  It may be that $x_{ij} \in \mathbb{R}$ or $x_{ij}$ may be a categorical value.  In both cases, consider a partition of the support of variable $j$ to be quantised $\mathcal{K}_j = \{ K^j_1, \dots, K^j_m \}$.  That is, $x_{ij} \in \bigcup_{k=1}^m K^j_k \ \forall\ i, j$ and $K^j_i \cap K^j_k = \varnothing \ \forall\ j, \forall\ i \ne k$.  There are at least two routes one may take to quantisation:

\begin{enumerate}
  \item $x_{ij}$ is encoded as an indicator $\tilde{x}_{ijk} \in \{0,1\} \ \forall\,k$, where for each continuous variable $j$, $\tilde{x}_{ijk} = 1 \iff x_{ij} \in K^j_k$ and $\tilde{x}_{ijl} = 0 \ \forall\,l \ne k$.  For example, a natural ordinal choice for $\mathcal{K}_j$ is the partition induced by the quintiles of that variable.
  \item If the partition also satisfies $y < z \iff y \in K^j_i, z \in K^j_k$ and $i<k \ \forall\ j$, then another option is to replace the value of $x_{ij}$ by a corresponding ordinal value, so that $\tilde{x}_{ij} = k \iff x_{ij} \in K^j_k$ so that the support becomes $\tilde{x}_{ij} \in \{ 1, 2, \dots, m_j \}$.
\end{enumerate}

Both approaches transform continuous, categorical or ordinal values to an encoding which can be represented directly in the message space of homomorphic schemes.  Note that for categorical or discrete variables in the design matrix, these procedures can be exact, whilst for continuous ones they may introduce an approximation.

Thus, for example a design matrix $X$ would map to two possible representations $\tilde{X}_1$ and $X_2$ corresponding to the different procedures above, with $X_{i1} \in \{0,1\}, X_{i2} \in \{1,2,3\}$ and $X_{i3} \in \mathbb{R}$:
\begin{align*}\label{eq:Xcomplex}
X = \left( 
\begin{array}{ccc}
{\cellcolor{red!20} 0} & {\cellcolor{green!20} 1} & {\cellcolor{blue!20} 1.7}  \\
{\cellcolor{red!50} 1} & {\cellcolor{green!50} 2} & {\cellcolor{blue!40} 1.9}  \\
{\cellcolor{red!80} 0} & {\cellcolor{green} 3}      & {\cellcolor{blue!60} 1.6}  \\
\vdots & \vdots & \vdots  \\
{\cellcolor{red!80} x_{i1}} & {\cellcolor{green} x_{i2}}      & {\cellcolor{blue!60} x_{i3}}
\end{array} 
\right) 
 &\to \tilde{X}_1 = \left(\begin{array}{ccccccccccccc} 
	{\cellcolor{red!20} 0}
  & {\cellcolor{green!20} 1}
  & {\cellcolor{green!20} 0}
  & {\cellcolor{green!20} 0}
  & {\cellcolor{blue!20} 0}
  & {\cellcolor{blue!20} 0}
  & {\cellcolor{blue!20} 1}
  & {\cellcolor{blue!20} 0}
  & {\cellcolor{blue!20} 0} \\
  	{\cellcolor{red!50} 1}
  & {\cellcolor{green!50} 0}
  & {\cellcolor{green!50} 1}
  & {\cellcolor{green!50} 0}
  & {\cellcolor{blue!40} 0}
  & {\cellcolor{blue!40} 0}
  & {\cellcolor{blue!40} 0}
  & {\cellcolor{blue!40} 0}
  & {\cellcolor{blue!40} 1} \\
  	{\cellcolor{red!80} 0}
  & {\cellcolor{green} 0}
  & {\cellcolor{green} 0}
  & {\cellcolor{green} 1}
  & {\cellcolor{blue!60} 1}
  & {\cellcolor{blue!60} 0}
  & {\cellcolor{blue!60} 0}
  & {\cellcolor{blue!60} 0}
  & {\cellcolor{blue!60} 0} \\
\vdots && \vdots  &&&& \vdots && \\
  	{\cellcolor{red!80} \tilde{x}_{i11}}
  & {\cellcolor{green} \tilde{x}_{i21}}
  & {\cellcolor{green} \tilde{x}_{i22}}
  & {\cellcolor{green} \tilde{x}_{i23}}
  & {\cellcolor{blue!60} \tilde{x}_{i31}}
  & {\cellcolor{blue!60} \tilde{x}_{i32}}
  & {\cellcolor{blue!60} \tilde{x}_{i33}}
  & {\cellcolor{blue!60} \tilde{x}_{i34}}
  & {\cellcolor{blue!60} \tilde{x}_{i35}}
  \end{array}\right) \\
\ \mbox{ or, under method 2 } &\to X_2 = \left( 
\begin{array}{ccc}
{\cellcolor{red!20} 0} & {\cellcolor{green!20} 1} & {\cellcolor{blue!20} 3}  \\
{\cellcolor{red!50} 1} & {\cellcolor{green!50} 2} & {\cellcolor{blue!40} 5}  \\
{\cellcolor{red!80} 0} & {\cellcolor{green} 3}      & {\cellcolor{blue!60} 1}  \\
\vdots & \vdots & \vdots  \\
{\cellcolor{red!80} \tilde{x}_{i1}} & {\cellcolor{green} \tilde{x}_{i2}} & {\cellcolor{blue!60} \tilde{x}_{i3}}
\end{array} 
\right)
\end{align*}

Recall from \S\ref{sec:limitations} and \citet*{Part1} that comparisons of equality cannot usually be made on encrypted content.  However, Method 1 can be seen to enable encrypted indicators for simple tests of equality, since comparisons simply become inner products:
\begin{equation}
  \sum_{\forall\,k} \tilde{x}_{i_1jk} \tilde{x}_{i_2jk} = 1 \iff \mbox{ obs }i_1 \mbox{ and } i_2 \mbox{ have equal quantised value on variable } j
\label{eq:comparison}
\end{equation}
otherwise the sum is zero.  In particular, note that this is a homogeneous polynomial of degree 2, requiring only 1 multiplication depth in the analysis.

Likewise, it is possible to evaluate an encrypted indicator for whether a value lies in a given range, because:
\begin{equation}
  \sum_{k \in K} \tilde{x}_{ijk} = 1 \iff \mbox{ obs } i \mbox{ has quantised value in the set } K
\label{eq:range}
\end{equation}
otherwise the sum is zero.

Conversely, Method 2 may be preferred in linear modelling situations which would then represent the change in $y$ for an incremental change in the quantised encoding, whereas in a linear modelling context Method 1 results in separate estimates of effect for each category of encoding.

Note that this is not a binary representation of the kind critiqued in section \ref{sec:limitations} --- here they are binary indicator \emph{values}, with an integer \emph{representation} in an integer \emph{space}.  Therefore, to count the number of indicators, for example, is simple addition, as opposed to the binary arithmetic described earlier.

In the next two sections we present the tailored statistical machine learning techniques developed specifically with the constraints of homomorphic encryption in mind.

\section{Extremely Random Forests}
\label{sec:RF}

Extremely or perfectly random forests \citep{Geurts06,Cutler01}  can exhibit competitive classification performance against their more traditional counterpart \citep{breiman2001random}. Forest methods combine many decision trees in an ensemble classifier and empirically often perform well on complex non-linear classification problems. Traditional random forests involve extensive comparison operations and evaluation of split quality at each level, operations which are either prohibitive or impossible to compute homomorphically in current schemes. However, we show that a tailored version of extremely or perfectly random forests can be computed fully encrypted, where both fitting and prediction are possible, with all operations performed in cipher text space. Moreover we highlight that the completely random nature of the methods allows for incremental learning and divide-and-conquer learning on large data, so that massive parallelism can be employed to ameliorate the high costs of encrypted computation.  In particular, this is demonstrated in a real $1,152$ core cluster example in \S\ref{sec:AWS}.

\subsection{Completely Random Forests (CRF)}

To begin, we assume the training data are encoded as in Method 1 (\S\ref{sec:Quantisation}) so that the comparison identities in \eqref{eq:comparison} and \eqref{eq:range} can be used. In overview, the most basic form of the proposed algorithm then proceeds as follows:
\begin{itemize}
\item[Step 1.]  Predictor variables at each level in a tree are chosen uniformly (``completely'') at random from a subset of the full predictor set.  Additionally, the split points are chosen uniformly (``completely'') at random from a set of potential split points.  Identity \eqref{eq:range} then provides an indicator variable for which branch a variable lies in, so that a product of such indicators provides an indicator for a full branch of a decision tree.  Then \eqref{eq:comparison} enables the pseudo-comparison involved in counting how many observations of a given class are in each leaf of the tree. 
\item[Step 2.] Step 1 is repeated for each tree in the forest independently, using a random subset of predictors per tree, so that many such trees are grown.  Each observation casts one vote per tree, according to the terminal leaf and class to which it belongs. Note that Step 2 can be performed in parallel as the trees are grown independently of one another.
\item [Step 3.] At prediction the same identities as in Step 2 can be used to create an indicator which picks out the appropriate vote from each tree, for each class.
\end{itemize}

The detailed algorithm is given in Appendix \ref{sec:CRFalg}.

This algorithm is referred to as a `completely random forest' since it takes the random growth of trees to a logical extreme with tree construction performed completely blindfold from the data.  This is in contrast to, for example, extremely random forests \citep{Geurts06} where optimisation takes place over a random selection of splits and variables, and tree growth can terminate upon observing node purity or underpopulation.  It is also different to perfectly random tree ensembles \citep{Cutler01}, where random split points are constructed between observations known to belong to different classes.  Neither of those approaches can be directly implemented within the constraints of fully homomorphic encryption.

The model returns an encrypted prediction, 
$$\hat{\nu}^\star_c \leftarrow {\textrm{CRF}}(\tilde{x}, \tilde{y}, \tilde{x}^\star)$$ 
as a count of the votes\footnote{The vote is the total number of training samples of category $c$ laying in the same root node as prediction point $x^\star$ across the trees.} for each class category, $c$, in message space $y_i \in \mathcal{C}$, with encrypted training data $\{\tilde{x}, \tilde{y}\}$ and encrypted test prediction point $\tilde{x}^\star$.  The user decrypts using the private encryption key $\hat{z}^\star_c = \Dec(k_s, \hat{\nu}^\star_c)$ and forms a predictive empirical `probability' as:
    \begin{equation} \hat{p}^\star_c = \frac{\hat{z}^\star_c}{\sum_{c=1}^{|\mathcal{C}|} \hat{z}^\star_c} \label{eq:SFpredprob}\end{equation}

\subsection{Cryptographic stochastic fraction estimate}
\label{sec:RFfrac}

In conventional forest algorithms  each tree gets a single prediction `vote' regardless of the number of training samples that were present in a leaf node for prediction. This is in contrast to the above, where due to encryption constraints the algorithm simply counts the total number of training samples from each category falling in the leaf node of the prediction point, summed across all trees. The difficulty in matching to convention is that converting the number of training samples in a category to the vote of the most probable category for each tree is not possible under current FHE schemes, and would need to be done through decryption.

To address this we propose a method of making an asymptotically consistent stochastic approximation to enable voting from each tree.  This is done by exploiting the fact that the adjustment required can be approximated via an appropriate encrypted Bernoulli process by sampling with replacement.  This stochastic adjustment can be computed entirely encrypted.

There are several approaches to estimating class probabilities from an ensemble of trees.  Perhaps most common is the average vote forest, which is not possible because comparisons between class votes to establish the maximum vote in a leaf are not possible.  An alternative is the relative class frequencies approach, which appears also to be beyond reach encrypted because of the need to perform division and representation of values in $(0,1)$.  An obvious solution to the representation issue as already discussed in \S\ref{sec:datarep} is to say:
\[ \mbox{votes for class }c\mbox{ in leaf }b\mbox{ of tree }t = \left\lfloor \frac{N \rho_{bc}^t}{\sum_c \rho_{bc}^t} \right\rceil \approx \rho_{bc}^t \left\lfloor \frac{N}{\sum_c \rho_{bc}^t} \right\rceil \]
where $\left\lfloor \cdot \right\rceil$ denotes rounding to the nearest integer, $N$ is the number of training observations, $\rho_{bc}^t$ counts the number of training samples of class $c$ laying in the $b$th terminal leaf node of the $t$th tree (see Appendix \ref{sec:CRFalg} for details), giving a scale $\{0, \dots, N\}$ which can be represented encrypted, albeit seemingly still not computed due to the division.

Note that $\sum_c \rho_{bc}^t \le N$, so that the reciprocal of the second term above lies in $(0,1)$ and can be treated as a probability, and recall that $X\sim\mbox{Geometric}(p) \ \implies \ \mathbb{E}[X] = p^{-1}$.  In other words, one can view an unbiased stochastic approximation to the fraction we require to be a draw from a Geometric distribution with probability $\frac{\sum_c \rho_{bc}^t}{N}$.

This transforms the problem from performing division to performing encrypted random number generation, where the distribution parameter involves division, which initially may seem worse.  However, observe that each $\rho_{bc}^t$ term arises from summing a binary vector from $\{0,1\}^N$:
\[ \rho^{t}_{bc} = \sum_{i=1}^N \eta_{ibc}^t \quad \mbox{ where } \quad \eta_{ibc}^t := \tilde{y}_{ic} I( \tilde{x}_i \in b ) \]
Consequently, exchanging the order of summation, $\sum_c \rho_{bc}^t$ can be treated as a sum of a binary vector length $N$:
\[ \sum_{c=1}^{|\mathcal{C}|} \rho_{bc}^t = \sum_{i=1}^N \eta_{ib}^t \quad \mbox{ where } \quad \eta_{ib}^t = \sum_{c=1}^{|\mathcal{C}|} \eta_{ibc}^t \]
where $|\mathcal{C}|$ is the number of classes.  In other words, the length $N$ vector $\eta_{ib}^t$ (for each $t, b$) is an encrypted sequence of 0's and 1's with precisely the correct number of 1's such that blind random sampling with replacement from the elements produces an (encrypted) Bernoulli process with success probability $\frac{\sum_c \rho_{bc}^t}{N}$.  Hereinafter, refer to this as simply $\eta_i$, the dependence on tree and leaf being implicit.

Thus, the objective is to sample a Geometric random variable encrypted, but at this stage it is only possible to generate the encrypted Bernoulli process underlying the desired Geometric distribution.  This finally shifts the problem to that of counting the number of leading zeros in an encrypted Bernoulli process: in other words, resample with replacement from $\eta_1, \dots, \eta_N$ a vector of length $M$, say, and without decrypting establish the number of leading zeros.

To achieve this it is possible to draw on an algorithm used in CPU hardware to determine the number of leading zeros in an IEEE floating point number, an operation required when renormalising the mantissa (the coefficient in scientific notation).  Let $\eta_1, \dots, \eta_M$ be a resampled vector and assume $M$ is a power of 2 (the reason being that this maximises the estimation accuracy for a fixed number of multiplications):
\begin{enumerate}
  \item For $l \in \{ 0, \dots, \log_2(M)-1 \}$:
    \begin{itemize}
      \item Set $\eta_i = \eta_i \lor \eta_{i-2^l} = \eta_i+\eta_{i-2^l}-\eta_i\eta_{i-2^l} \quad \forall\,2^l+1\le i \le M$
    \end{itemize}
  \item The number of leading zeros is $M-\sum_{i=1}^M \eta_i$
\end{enumerate}
In summary, this corresponds to increasing power of 2 bit-shifts which are then OR'd with itself, all of which can be computed encrypted.

Thus, an approximately unbiased (for large enough $M$) encrypted estimator of the desired fraction, $\left\lfloor \frac{N}{\sum_c \rho_{bc}^t} \right\rceil$, is $M-\sum_{i=1}^M \eta_i+1$ upon termination of the above algorithm. It is important to note that the multiplicative depth required for this algorithm is $M$, and recall that multiplicative depth is restricted under current FHE schemes \citep*{Part1} if expensive cipher text refreshing is to be avoided.  Hence, in practise typically the resample size will be restricted to a small value like $M=32$ even for large $N$ datasets.  However, this is desirable: it enables some shrinkage to take place by placing an upper bound of $M$ on the fraction estimate.  Thus, for a choice of $M=32$, terminal leaves will in expectation have the correct adjustment if at least $\frac{1}{32}$ of the training data are in that decision path of the tree.  For example, in a training data set of size $N=1000$, the stochastic fraction is correct in expectation for all leaves containing at least $\frac{1000}{32} = 31$ observations --- fewer observations and the leaf votes will undergo shrinkage in expectation.

Note in particular that CRFs are inherently discrete and the probability of regrowing exactly the same tree twice is not measure zero, so that asymptotically the same tree will be regrown infinitely often with probability 1.  If the encrypted stochastic fraction is recomputed in each new tree then asymptotically the correct adjustment will be made.

\subsection{Further implementation issues}

We highlight a couple of additional implementation issues that are important for the practical machine learning of completely random forests on FHE data.

\subsubsection{Calibration}

The first point to note is that there is no calibration of the trees, or indeed the forest.  Consequently there should be no presumption that $\argmax_c \{p_c^\star : c=1,\dots,|\mathcal{C}|\}$ provides the ``best'' prediction for class $c$ under unequal misclassification loss.  As such, the traditional training and testing setup is crucially important in order to select optimal decision boundaries according to whatever criteria are relevant to the subject matter of the problem, such as false positive and negative rates.  This is the only step which must be performed unencrypted: the responses of the test set must be visible, though note that the predictors need not since step 3 for prediction is computed homomorphically.

\subsubsection{Incremental and parallel computation}
\label{sec:CRFparallel}

One key advantage of CRFs is that learning is incremental as new data become available: there is no need to recompute the entire fit as there is no optimisation step, so that once used encrypted data can be archived at lower cost and moreover adding new observations has linear growth in computational cost.

Indeed the whole algorithm is embarrassingly parallel both in the number of trees and the number of observations.  One can independently compute the trees and data can be split into shards whereby $\rho_{bc}^t$ is computed for each shard separately using the same seed in the random number generator for growing trees and then simply additively combined afterwards (a comparatively cheap operation). This is highlighted in a real-world large scale example in \S\ref{sec:AWS}.

\subsubsection{Theoretical parameter requirements for \citet{Fan12}}

For a discussion of practical requirements for the parameter selection in homomorphic encryption schemes, see Appendix \ref{sec:pars}.

\vspace{12pt}

The CRF fitting, prediction and forest combination are all implemented in the open source R package \texttt{EncryptedStats} \citep{ESpkg} and can be run on unencrypted data as well as data encrypted using the \texttt{HomomophicEncryption} \citep{HEpkg} package.  These are briefly described in Appendix \ref{sec:APDX_implementation}.

The next section introduces the second novel method tailored for FHE.

\FloatBarrier

\section{Na\"{i}ve Bayes Classifiers}
\label{sec:NaiveBayes}

The Na\"{i}ve Bayes (NB) classifier is a popular generative classification algorithm that models the joint probability of predictor variables independently for each response class, and then uses Bayes rule and an independence assumption among predictors to construct a simple classifier \citep[p.210]{NgJordan02,hastie2009elements}. The advantages and disadvantages have been extensively described in the literature, for example \citep{RennieETAL03}. Although the independence assumption underlying NB is often violated, the linear growth in complexity for large number of predictors and the simple closed-form expressions for the decision rules make the approach attractive in ``big-data'' situations. Moreover as highlighted by \cite{DomingosPazzani97}  there is an important distinction between classification accuracy (predicting the correct class) and accurately estimating the class probability,  and hence NB can perform well in classification error rate even when the independence assumption is violated by a wide margin.
Essentially, although it produces biased probability estimates, this does not necessarily translate into a high classification error \citep{HandYu01}.
Consequently, NB remains a well established and popular method.

\subsection*{The Na\"{i}ve Bayes framework}

Consider the binary classification problem with a set of $P$ predictors $x = \{ x_{j} \}_{j \in \mathbb{N}_{1:P}} \in \mathcal{X}$ and let
$y \in \{0,1\}$. 
The NB classifier uses Bayes theorem,
\begin{equation}
\mathbb{P}(y|x) \propto \mathbb{P}(x|y) \mathbb{P}(y),
\end{equation}
for prediction coupled with an independence assumption, 
\begin{equation}
\mathbb{P}(x|y) = \prod_{j=1}^P \mathbb{P}(x_{j} | y),
\end{equation}
which embodies a compromise between accuracy and tractability, to obtain a conditional class probability.
This allows NB to separately model the conditional distributions of predictor variables, $\{\mathbb{P}(x|y=1), \mathbb{P}(x|y=0)\}$,  and then construct the prediction probability via Bayes theorem,
\begin{equation} \label{eq:NBprobY1}
\mathbb{P}(y=1|x) 
    = \frac{    \mathbb{P}(x|y=1) \mathbb{P}(y=1) 
            }{ 
                \mathbb{P}(x|y=1) \mathbb{P}(y=1) 
                + \mathbb{P}(x|y=0) \mathbb{P}(y=0)
            }.
\end{equation}
The most popular forms for the distributions $\mathbb{P}(x_{j} | y)$ are multinomial for categorical predictors and Gaussian for continuous predictors. As shown in the Appendix \ref{sec:MNB}, it is possible to work with multinomial distributions directly in cipher text space, albeit at a multiplicative depth of $3P-1$, but the Gaussian distributions lay outside of FHE.  In the next subsection we propose a tailored semi-parametric NB model that is amenable to cipher text computation.  Crucially this novel method scales to an arbitrary number of predictors at no additional multiplicative depth, making it well suited to encrypted computation.

\subsection{Semi-parametric Na\"{i}ve Bayes}
\label{sec:NB_SNB}
The NB classifier solves the classification task using a generative approach, i.e., by modelling the distribution of the predictors \citep{NgJordan02}.  However, distributions such as the Gaussian cannot be directly implemented within an FHE as they involve division and exponentiation operators as well as continuous values.  
Here, we show that it is possible to model the decision boundary between the two response classes more explicitly --- without a parametric model for the distributions of the predictors --- while still remaining in the NB framework.
As will become clear, this corresponds to a discriminative approach to classification where the decision boundary of the conditional class probabilities is modelled semi-parametrically.

To begin, note that the expression for the log-odds prediction from NB can be rearranged to give
\begin{equation}
\begin{aligned}
\log \left( \frac{\mathbb{P}(y=1 | x)}{\mathbb{P}(y=0 | x)} \right) 
& = \log \left( 
	\frac{ \mathbb{P}(y=1) }{ \mathbb{P}(y=0) }  
	\prod_{j=1}^{P} 
	\frac{ \mathbb{P}(x_j | y=1) }{ \mathbb{P}(x_j | y=0) } \right)  \\ 
& = \log \left(  \left[\frac{\mathbb{P}(y=0)}{\mathbb{P}(y=1)}\right]^{(P-1)} \prod_{j=1}^{P} \left[ \frac{\mathbb{P}(x_j | y=1)}{\mathbb{P}(x_j | y=0)} \cdot  \frac{\mathbb{P}(y=1)}{\mathbb{P}(y=0)} \right] \right)  \\
 & = \log \left( \left[\frac{\mathbb{P}(y=0)}{\mathbb{P}(y=1)}\right]^{(P-1)} \prod_{j=1}^{P} \frac{\mathbb{P}(y =1 | x_j)}{\mathbb{P}(y=0 | x_j)} \right)  \\
 & = (P-1)\log \left(\frac{\mathbb{P}(y=0)}{\mathbb{P}(y=1)} \right) + \sum_{j=1}^{P} \log \left( \frac{\mathbb{P}(y =1 | x_j)}{\mathbb{P}(y=0 | x_j)} \right).
\end{aligned}
\end{equation}
Using this identity, we now propose to model the decision boundary $\mathbb{P}(y | x_j)$ directly using the linear logistic form,  rather than parameterise $\mathbb{P}(x_j | y)$ via a distribution function. That is we assume,
\begin{equation}\label{eq:SNB_f}
\log \left( \frac{\mathbb{P}(y =1 | x_j)}{\mathbb{P}(y=0 | x_j)} \right) = f(x_{j}; \alpha_{j}, \beta_{j})
\qquad\text{and}\qquad 
\frac{\mathbb{P}(y=0)}{\mathbb{P}(y=1)} = \theta
\end{equation}
where, in this work, $f(x_{j}; \alpha_{j}, \beta_{j})$ is taken to be a linear predictor of the form $\alpha_{j} + \beta_{j} x_{j}$.  The independence structure means that each term can be optimised independently by way of an approximation to logistic regression, amenable to homomorphic computation (presented below; \S\ref{sec:stats_glm}), since the standard iteratively reweighted least squares fitting procedure is not computable under the restrictions of homomorphic encryption.  Optimisation of $\theta$ is done independently of \mbox{$\{\alpha_{j}, \beta_{j}\}$},
\begin{equation} \label{eq:theta}
\hat{\theta} = 
\frac{\sum_{i=1}^{N} y_{i}}{N - \sum_{i=1}^{N} y_{i}}.
\end{equation}
The estimated log-odds are then
\begin{equation} \label{eq:SNB_predict_a}
\hat{\psi}( x; \hat{\theta}, \hat{\alpha}, \hat{\beta} ) 
= \log \left( \frac{\mathbb{P}(y=1 | x; \hat{\theta}, \hat{\alpha}, \hat{\beta} ) }{\mathbb{P}(y=0 | x; \hat{\theta},  \hat{\alpha}, \hat{\beta} )} \right)  
= (P-1) \log(\hat{\theta}) +  \sum_{j=1}^{P} f ( x_j ; \hat{\theta}, \hat{\alpha}_{j}, \hat{\beta}_{j} )
\end{equation}
or, equivalently, in terms of conditional class probabilities
\begin{equation} \label{eq:SNB_predict_b}
\mathbb{P}(y=1 | x; \hat{\theta}, \hat{\alpha}, \hat{\beta}) = \frac{ 1 }{ 1 + \exp\{ - \hat{\psi}( x; \hat{\theta}, \hat{\alpha}, \hat{\beta} )  \}}.
\end{equation}
Hence equation \eqref{eq:SNB_predict_b} can be computed after decryption from the factors (numerator/denominator) that comprise $\{ \hat{\theta}, \hat{\alpha} , \hat{\beta} \}$ to form the conditional class predictions.

This defines a semi-parametric Na\"{i}ve Bayes (SNB) classification model that assumes a logistic form for the decision boundary in each predictor variable, and where each logistic regression involves at most two parameters. 
As we will show in the next section, in this setting there is a suitable approximation to the maximum likelihood estimates which are traditionally computed using the iteratively reweighted least squares algorithm.

\subsection{Logistic Regression}
\label{sec:stats_glm}

The SNB algorithm proposed in the previous section requires a homomorphic implementation of simple logistic regression involving an intercept and a single slope parameter.
In this section an approximation to logistic regression based on the first iteration of the Iteratively Reweighted Least Squares (IRLS) algorithm is proposed and some of its theoretical properties analysed. Apart from its use in SNB, the approximation to logistic regression described in this section also stands on its own as a classification method amenable to homomorphic computation.

\subsubsection{First step of Iteratively Reweighted Least Squares}

Optimisation in logistic regression is typically achieved via IRLS based on Newton--Raphson iterations.
Starting from an initial guess $\beta^{(0)}$, the first step involves updating the auxiliary variables
\begin{equation}\begin{aligned}
	\eta_{i}^{(0)}	&= X_{i\smallbullet}\beta^{(0)}
 \\
	\mu_{i}^{(0)} &= \frac{\exp\left( \eta_{i}^{(0)} \right) }{ 1 + \exp\left( \eta_{i}^{(0)} \right) } \\
	w_{ii}^{(0)} &= \mu_{i}^{(0)} \left( 1 - \mu_{i}^{(0)} \right) \\
	z_{i}^{(0)}	&= \eta_{i}^{(0)} +\left( w_{ii}^{(0)} \right)^{-1} \left( y_{i} - \mu_{i}^{(0)} \right) 
\end{aligned}\end{equation}
where $X_{i\smallbullet}$ denotes the $i$th row of $X$.
By starting with an initial value $\beta^{(0)} = (0,\dots,0)^{T}$ the initialisation step of the algorithm is simplified: for all $i \in \mathbb{N}_{1:N}$ we have 
$\eta_{i}^{(0)}=0$, 
$\mu_{i}^{(0)}=1/2$, 
$w_{i,i}^{(0)}=1/4$ 
and 
$z_{i}^{(0)} = 4  y_{i} - 2$. 
In the second step, the parameter estimates of $\beta$ are updated using generalised least squares
\begin{equation} \label{eq:betaIRLSupdate}
	\beta^{(1)} = \left( X^{T}W^{(0)}X \right)^{-1} X^{T}W^{(0)}z^{(0)}
\end{equation}
and these two steps are repeated until convergence is achieved.

In what follows, only the first-iteration update of $\beta$ is considered, because it provides an approximation to logistic regression and, most importantly, one which is computationally feasible under homomorphic encryption; this will be termed the \textit{one-step approximation}.
Note that implementation of the full IRLS algorithm is infeasible under FHE because the weights $w_{ii}$ can not be updated as this requires the evaluation of non-polynomial functions.

The particular form of these (\textit{paired}) one-step approximating equations for an intercept and a single slope parameter is shown in the following section.
We show an additional simplification (so-called \textit{unpaired} estimates) which improves encrypted computational characteristics further at the cost of greater approximation.

\subsubsection{Paired, one-step approximation to simple logistic regression}	\label{sec:glm_1step_2D}

In the SNB model the conditional log-odds are optimised for each variable separately, following the independence assumption. 
In this case the the design matrix $X$ contains a single column of 1's  for the intercept and a single predictor variable column, and the first step of the IRLS leads to
\begin{equation}
    \{ \hat{\alpha}_{j}, \hat{\beta}_{j} \} = \left\{ a_{j} / d_{j}, b_{j} / d_{j} \right\}, \quad \forall j \in \mathbb{N}_{1:P}
\end{equation}
and where
\begin{align}
    a_{j} &= \left( \sum_{i=1}^{N} x_{ij}^{2} \right) \left( \sum_{i=1}^{N} z_{i}^{(0)} \right) - \left( \sum_{i=1}^{N} x_{ij} \right) \left( \sum_{i=1}^{N} x_{ij}z_{i}^{(0)} \right)  \\
    b_{j} &= N \sum_{i=1}^{N} x_{ij}z_{i}^{(0)}  - \left( \sum_{i=1}^{N} x_{ij} \right) \left( \sum_{i=1}^{N} z_{i}^{(0)} \right)  \\
    d_{j} &= N \sum_{i=1}^{N} x_{ij}^{2} - \left( \sum_{i=1}^{N} x_{ij} \right)^{2} \label{eq:1step_d}
\end{align}
To use the one-step approximation in combination with the SNB classifier detailed above (\S\ref{sec:NB_SNB}), the quantity 
$\{ \hat{\alpha}_{j} + \hat{\beta}_{j} x_{j}^{\star} \}$ 
--- required for the classification of a new observation \mbox{$x^{\star} = \{ x_{j} \}_{j \in \mathbb{N}_{1:P}}$} --- can then be estimated as $e_{j}/d_{j}$, where 
$e_{j} = a_{j} + b_{j} x_{j}^{\star}$.

This corresponds to a standard, \textit{paired} optimisation strategy, that is, to optimise intercept and slope jointly, using an approximation targeting
\begin{equation} \label{eq:NBopt_paired}
\underset{\alpha_{j}, \beta_{j}}{\arg\sup}\; f(x_{j}; \alpha_{j}, \beta_{j}), \quad \forall j \in \mathbb{N}_{1:P}.
\end{equation}

An alternative to this approach would be to optimise intercept and slope independently (i.e., in an \textit{unpaired} fashion), that is, targeting
\begin{equation} \label{eq:NBopt_unpaired}
\underset{\alpha_{j}}{\arg\sup}\; f(x_{j}; \alpha_{j}, \beta_{j}=0) 
\quad \text{and} \quad
\underset{\beta_{j}}{\arg\sup}\; f(x_{j}; \alpha_{j}=0, \beta_{j})
, \quad \forall j \in \mathbb{N}_{1:P}
\end{equation}
in which case all $\alpha_{j}$ take the same value and, therefore, this is equivalent to estimating all $\beta_{j}$ independently and including a global intercept, $P \alpha_j$ for some $j$.

We will see that this is computationally appealing  due to the simplified estimating equations when using the same one-step approximation.

\subsubsection{Unpaired, one-step approximation to simple logistic regression}	\label{sec:glm_1step_1D}

In the case of unpaired estimation (or in the absence of an intercept term) the estimating equations for $\alpha_{j}$ and $\beta_{j}$ are simpler. 
To distinguish between unpaired and paired estimates we denote the unpaired by $\tilde{\alpha}_{j}$ and $\tilde{\beta}_{j}$, respectively.
All $\tilde{\alpha}_{j}$ have a common form
\begin{equation}
\tilde{\alpha}_{j} 
= \frac{1}{N} \sum_{i=1}^{N} z_{i}^{(0)}
\end{equation}
while $\tilde{\beta}_{j}$ have the form
\begin{equation}\label{eq:bet1step_x}
\tilde{\beta}_{j} = \frac{ \sum_{i=1}^{N} x_{ij} z_{i}^{(0)} } { \sum_{i=1}^{N} x_{ij}^{2} }.
\end{equation}

Note that this unpaired formulation also arises in the case of centred predictors, \mbox{$\mathbb{E}[x_j]=0$}, so that the unpaired approach is completely equivalent to the paired approach and introduces no additional approximation if it is possible for the data to be centred prior to encryption.  Note that this is trivially achievable for ordinal quintile data by representing each quintile by $\{-2,-1,0,1,2\}$, rather than $\{1,2,3,4,5\}$.  Consequently, where it is possible to centre the data it should be done in order to take advantage of these computational benefits at no approximation cost.

Furthermore, when the data are represented as in section \ref{sec:Quantisation}, this can be rewritten as follows.
Define the following auxiliary variables
\begin{equation}
n_{jk}^{[1]}	= \sum_{i:y_{i}=1} \tilde{x}_{ijk} 
				= \sum_{i:\tilde{x}_{ijk}=1} y_{i}	,\qquad\quad
n_{jk} 			= \sum_{i=1}^{N} \tilde{x}_{ijk}	,\qquad\quad
n_{jk}^{[0]} 	= n_{jk} - n_{jk}^{[1]}
\end{equation}
In words, $n_{jk}^{[c]}$ counts the number of observation in the $k$th bin of the $j$th predictor ---or, equivalently, the number of elements of $X_{\smallbullet j}$ (the $j$th column of $X$) which are equal to $k$---and for which the corresponding response is equal to $c$, for $c \in \{0,1\}$. In this case, Equation~\eqref{eq:bet1step_x} becomes
\begin{equation}\label{eq:bet1step_xtilde}
\tilde{\beta}_{j} 
= 2 \left( \sum_{k=1}^{M_{j}} k^{2} \sum_{i=1}^{N} \tilde{x}_{ijk}\right)^{-1} \sum_{k=1}^{M_{j}} k \left( n_{jk}^{[1]} - n_{jk}^{[0]}\right)
\end{equation}
so that for a binary predictor $x_{ij} \in \{0,1\}$ we find
\begin{equation} \label{eq:betatilde}
	\tilde{\beta}_j = \frac{2}{n} \left( n_j^{[1]} - n_j^{[0]} \right).
\end{equation}
where $n=\sum_i x_{ij}$.
This simple expression aids in the derivation of some theoretical results regarding the bias of the estimator.

\begin{figure}[!t]
\vspace{-0.4cm}
  \begin{center}
    \includegraphics[width=0.5\textwidth]{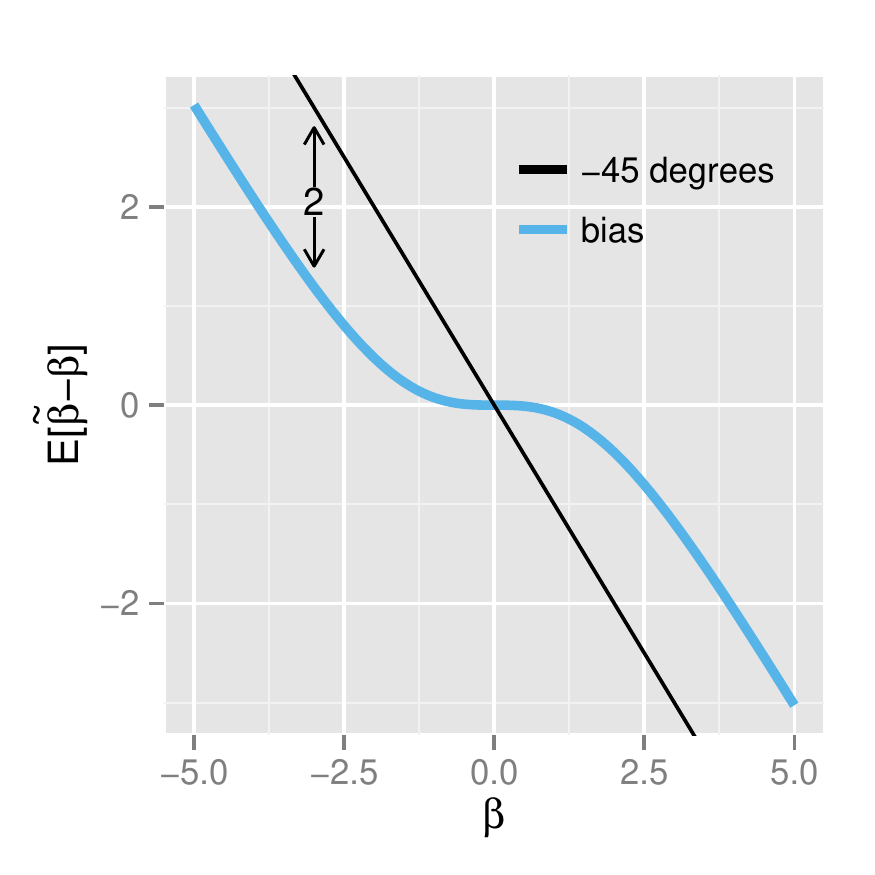}
  \end{center}
  \vspace{-0.5cm}
  \caption{Shrinkage in $\tilde{\beta}$. Because the estimator $\tilde{\beta}$ is bounded between -2 and 2, the shrunk value is always equal to 2 for absolute values of $\beta$ greater than $\approx 2.5$.}
  \label{fig:biasbeta1d}
\end{figure}

\subsubsection*{Shrinkage of $\boldsymbol{\tilde{\beta}}$}
Let $\beta$ denote the true parameter and $\tilde{\beta}$ denote the one-step IRLS estimate from Equation \eqref{eq:betatilde}.
The one-step ``early stopping'' shrinks or ``regularises'' the estimate towards the origin as, 
\begin{equation}
	\mathbb{E}[\tilde{\beta}-\beta]
		= \mathbb{E} \left[ \frac{ 2 \left( n^{[1]} - n^{[0]} \right) }{ n } - \beta \right]
		= - 2 + 4 \left(\frac{e^{\beta}}{1+e^{\beta}}\right) - \beta. \label{eq:1step_bias} 
\end{equation}
The shrinkage, as a function of the true parameter $\beta$, is shown in Figure~\ref{fig:biasbeta1d}; it is negligible when $\beta$ is small, but increases linearly with it (in magnitude) for $\beta$ outside the interval $(-2,2)$.
The reason for this is clear from the formula in Equation \eqref{eq:betatilde}: the range of $\tilde{\beta}$, the one-step estimate being $(-2,2)$.

In particular, note that this shrinkage is a highly desirable property in light of the independence assumptions made in SNB.  Indeed, the one-step procedure empirically outperforms the full convergence IRLS when predictors are highly correlated and moreover does not significantly underperform otherwise (tests were performed on all datasets to be presented in \S\ref{sec:results}).  Therefore the one-step method is not only a computational necessity for computing encrypted, but offers potential improvements in performance against the standard algorithm.

\subsubsection*{Generalisation error}

\begin{figure}[!t]
\captionsetup[subfigure]{belowskip=-1cm}
\centering
\subfigure{
	\includegraphics[width=0.5\textwidth]{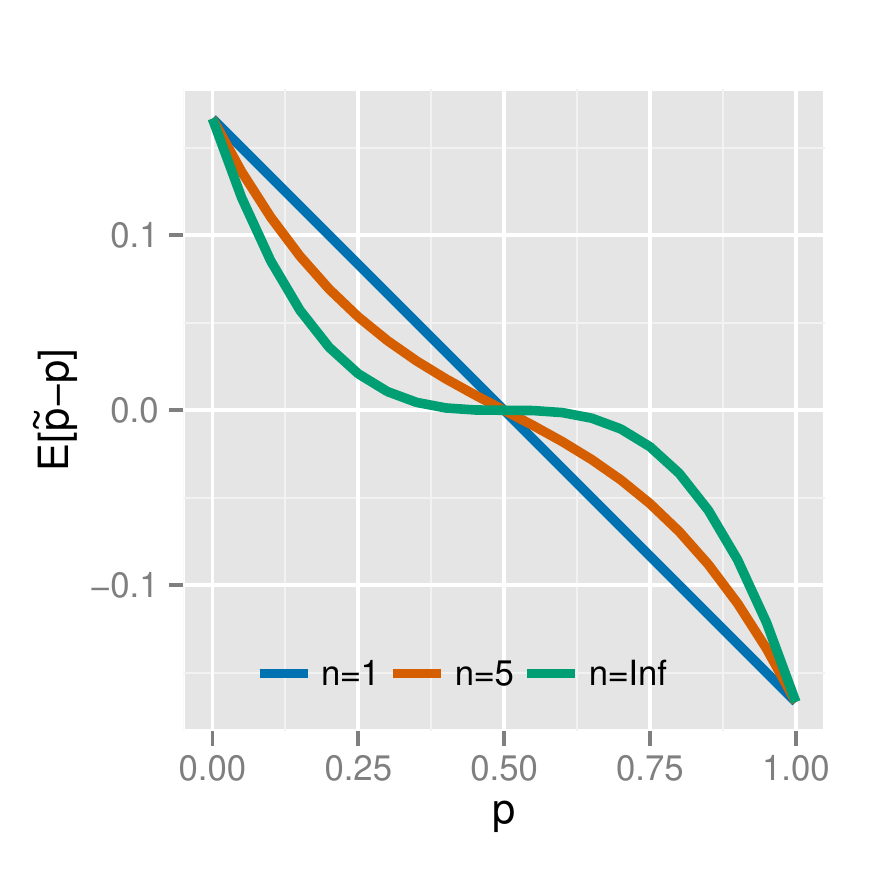}
	\label{fig:bias_beta_1d:a}
	}%
\subfigure{
	\includegraphics[width=0.5\textwidth]{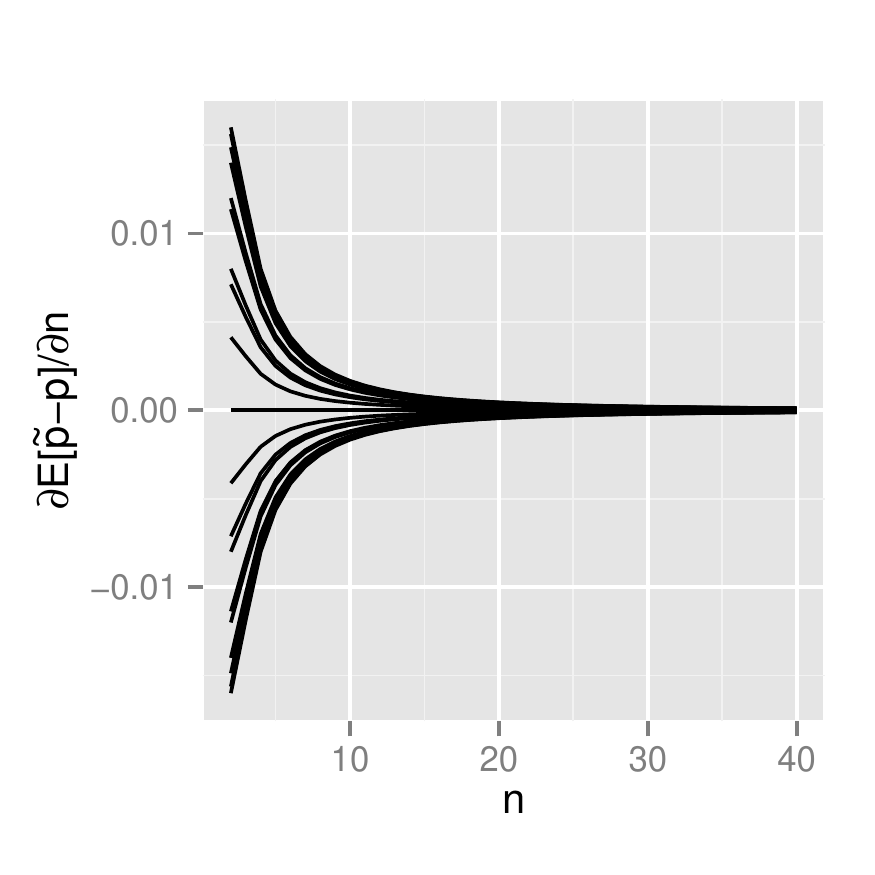}
	\label{fig:bias_beta_1d:b}
	}
\vspace{-0.5cm}
\caption{\textbf{(Left)} For all $p$, the generalisation error shrinks monotonically with increased $n$ and converges to an asymptotic curve---in green; \textbf{(Right)} The rate at which the generalisation error curves approach the asymptotic curve is decreasing with $n$ for all values of $p$ (several shown here) and stabilises at around $n=20$.} 
\label{fig:bias_beta_1d}  
\end{figure}

Define $p = \mathbb{P}[y=1 | x = 1, \beta]$ and $\tilde{p} = \mathbb{P}[y=1 | x = 1, \tilde{\beta}]$. Then, the generalisation error for $x=1$ can be written as 
\begin{equation}
	\mathbb{E}[\tilde{p} - p]
		= \mathbb{E} \left[ \frac{e^{\tilde{\beta}}}{1+e^{\tilde{\beta}}} - p \right]  
\end{equation}
and approximated by the polynomial
\begin{equation}
\mathbb{E}[\tilde{p} - p]
	\approx \mathbb{E} \left[ \frac{1}{2} + \frac{1}{4} \cdot \tilde{\beta} - \frac{1}{48} \cdot \tilde{\beta}^{3} \right] - p.
\end{equation} 
Using the moment generating function for the Binomial distribution one can compute the higher-order expectations and thus arrive at a formula which depends only on $n$ and $p$,
\begin{equation}
\begin{aligned}
\mathbb{E}[\tilde{p} - p]
	&\approx -\frac{1}{6n^{2}} \left( 8n^{2}p^{3} - 12n^{2}p^{2} + 6n^{2}p - n^{2} - 24np^{3} \right.  \\
	& \qquad\qquad\quad \left.  + 36np^{2} - 12np + 16p^{3} - 24p^{2} + 8p  \right).
\end{aligned}
\end{equation}
The expression is dominated by $p$ in the sense that even for very small values of $n$, the generalisation error is close (according to our approximation) to the one obtained asymptotically,
\begin{equation}
	\lim_{n\to\infty} \mathbb{E}[\tilde{p} - p] = -\frac{1}{6}(2p-1)^3.
\end{equation}

Figure~\ref{fig:bias_beta_1d} (left) shows how the value of $n$ affects the generalisation error; 
and Figure~\ref{fig:bias_beta_1d} (right) shows the speed at which the generalisation error converges to the one given by the (approximate) asymptotic curve, for several values of $p$. 

The generalisation error results highlight that while the estimates of the true class conditional probabilities may be unstable, the classification error rate achieved by SNB may be low as the classifier only has to get the prediction on the correct side of the decision boundary.

\subsubsection{Theoretical parameter requirements for \citet{Fan12}}

For a discussion of practical requirements for the parameter selection in homomorphic encryption schemes, see Appendix \ref{sec:pars}.

\vspace{12pt}

The SNB fitting and prediction for both paired and unpaired approximations are implemented in the open source R package \texttt{EncryptedStats} \citep{ESpkg} and can be run on unencrypted data as well as data encrypted using the \texttt{HomomophicEncryption} \citep{HEpkg} package.  These are briefly described in Appendix \ref{sec:APDX_implementation}.

In the next section, the two new machine learning techniques tailored for homomorphic encryption which have been presented are empirically tested on a range of data sets and a real example using a cluster of servers to fit a completely random forest is described.

\section{Results}
\label{sec:results}

In this section we apply the encrypted machine learning methods  presented in Sections \ref{sec:RF} and \ref{sec:NaiveBayes}  to a number of benchmark learning tasks.

\subsection{Classification performance}

We tested the methods on 20 data sets of varying type and dimension from the UCI machine learning data repository \citep{Lichman13}, each of which is described in Appendix \ref{sec:datasets}.  For the purposes of achieving many test replicates, the results in this subsection were generated from unencrypted runs (the code paths in the \texttt{EncryptedStats} package for unencrypted and encrypted values are identical), with checks to ensure that the unencrypted and encrypted versions give the same results. Runtime performance for encrypted versions are given below.

\begin{figure}[!ht]
\captionsetup[subfigure]{belowskip=-1cm}
\centering
\includegraphics[width=\textwidth]{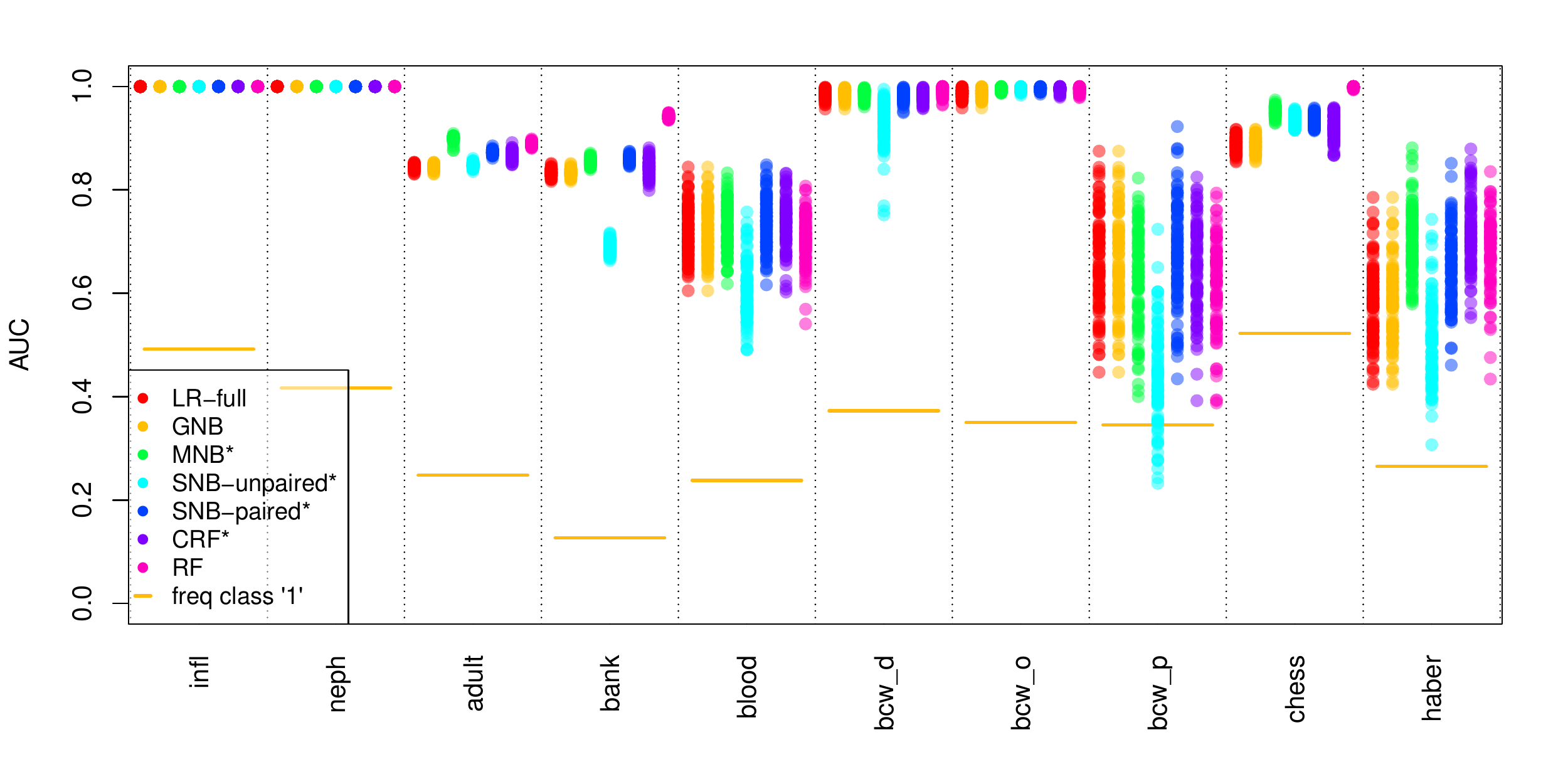}

\vspace{-0.7cm}
\includegraphics[width=\textwidth]{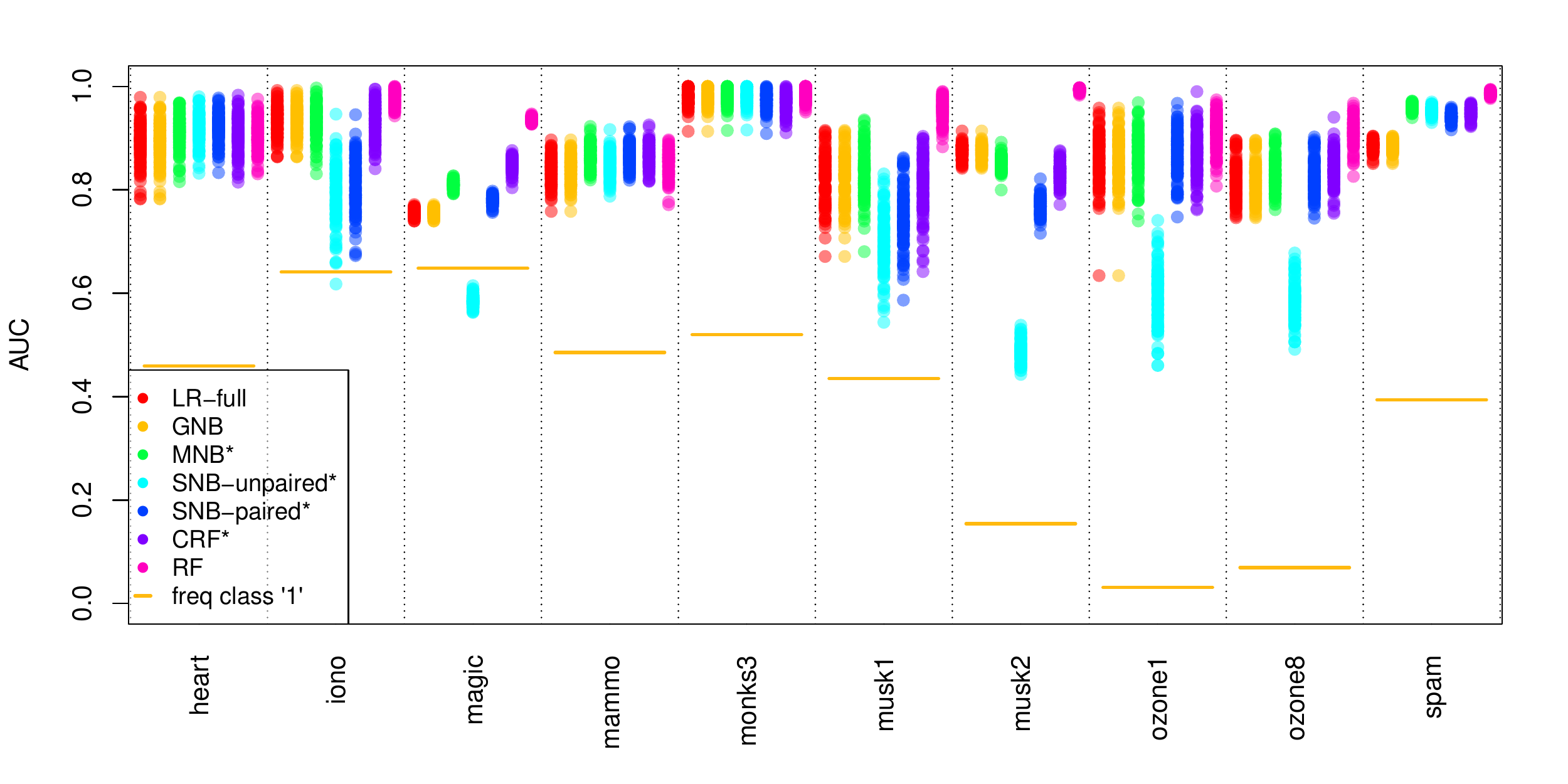}

\vspace{-0.4cm}
\caption{Performance of various methods. For each model and dataset, the AUC for 100 stratified randomisations of the training and testing sets; the horizontal lines represent the frequency of class $y=1$; an asterisk indicates the method can be computed encrypted.}
\label{fig:performance}
\end{figure}

Figure \ref{fig:performance} shows the comparison of these novel methods with each other as well as with their traditional counterparts.  The traditional methods included are full logistic regression (\texttt{LR-full}), Gaussian na\"{i}ve Bayes (\texttt{GNB}) and random forests (\texttt{RF}), none of which can be computed within the constraints of homomorphic encryption.  The methods of this paper included are, completely random  forests (\texttt{CRF}), paired (\texttt{SNB-paired}) and unpaired (\texttt{SNB-unpaired}) semi-parametric na\"{i}ve Bayes, and multinomial na\"{i}ve Bayes (\texttt{MNB}).  The CRFs are all 100 trees grown 3 levels deep, including stochastic fraction estimate ($M=8$).

The performance measure used is the area under the ROC curve (AUC, ranging from 0 to 1).  For each model and dataset the algorithms were run with the same 100 stratified randomisations of the training and testing sets (split in the proportion 80\%/20\%, respectively), so that each point on the graph represents the AUC for one train/test split and one method.

The first two data sets (\texttt{infl} and \texttt{neph}) are very easy classification problems and the new techniques match the traditional techniques perfectly in this setting, keeping almost uniformly good pace in the other rather easy data sets (\texttt{bcw\_d}, \texttt{bcw\_o}, \texttt{monks3}), the unpaired SNB being the exception on \texttt{bcw\_d}.

Unsurprisingly, the traditional random forest tends to perform best in the more challenging data sets \citep{fernandez2014we}, though only in 4 of the data sets is it clearly outperforming all the other methods by the AUC metric.  Indeed, in the most challenging data sets (\texttt{blood}, \texttt{bcw\_p} and \texttt{haber}) the new methods proposed in this work exhibit slightly better average performance than their counterparts.  

The results of the unpaired SNB were performed without centring (which would be equivalent to paired) and affirms the observation in the previous section that centring or paired computation is always to be preferred where available.

The SNB method with IRLS run to convergence is not presented in the figure: as alluded to in the previous section, the natural shrinkage of the one-step estimator meant it performed equally well in most situations and in the \texttt{chess} and \texttt{musk1} cases the average ratio of full convergence over one-step AUCs was 0.76 and 0.67 respectively.

As an aside, the \emph{unencrypted} version of these new methods have good computational properties which will scale to massive data sets, because the most complex operations involved are addition and multiplication which modern CPUs can evaluate with a few clock cycles latency.  In addition, in all cases even these simple operations can be performed in parallel and map directly to CPU vector instructions.

\subsection{Timings and memory use}

All the encrypted methods presented in this work can be implemented to scale reasonably linearly in the data set size, so performance numbers are provided per 100 observations and per predictor (for logistic regression and semi-parametric na\"{i}ve Bayes) or per tree (for completely random forests).

For reproducibility the timings were measured on a c4.8xlarge compute cluster Amazon EC2 instance.  This corresponds to 36 cores of an Intel Xeon E5-2666 v3 (Haswell) CPU clocked at 2.9GHz.  Table \ref{tab:times} shows the relevant timings using the \texttt{EncryptedStats} and \texttt{HomomorphicEncryption} packages.

\begin{table}[!h]
\captionsetup{width=0.8\textwidth}
\caption{Approximate running times (in seconds, on a c4.8xlarge EC2 instance) per 100 observations.  SNB are per predictor; CRF is per tree, $L$ is the tree depth grown.  (An `encrypted value' is a single integer value encrypted, so for example using quintiles each $x_{ij}$ is stored as 5 encrypted values.) \hfill{ }}\vspace{-0.6cm} 
\label{tab:times}
\begin{small}
\begin{center}
\begin{tabular}{p{9em}ccccccccccc}
\hline\hline 
& \multicolumn{4}{c}{\textbf{Model}} \\
& SNB-paired & CRF, $L=1$ & CRF, $L=2$ & CRF, $L=3$ \\
\cline{1-5}
Fitting & 18.0 & 12.5 & 45.2 & 347 \\
Prediction & 7.8 & 15.1 & 48.3 & 353 \\
\hline
Approx memory per encrypted value
& 154KB & 128KB & 528KB & $1,672$KB \\
\hline
\end{tabular}
\end{center}
\end{small}
\end{table}

Note that the CRF does \emph{not} scale linearly in the depth of the tree grown, not only because of the non-linear growth in computational complexity for trees but also because as $L$ increases so the parameters used for the \citet{Fan12} scheme have to be increased in such a way that raw performance of encrypted operations drops.  This drop is due to the increasing coefficient size and polynomial degree of the cipher text space.  See Appendix \ref{sec:pars} for a discussion of the impact of tree parameters on the encryption algorithm.

\subsection{Forest parameter choices}

The completely random forest has a few choices which can tune the performance: number of trees $T$, depth of trees to build $L$ and whether to use stochastic fractions (and to what upper estimate, $M$).  An empirical examination of these now follows.

\subsubsection{Forest trees and depth}
\label{sec:RF_T_LPerf}

We explored the effect of varying the number and depth of the trees used in the algorithm, by varying $T$ from 10 to 1000 and $L$ from 1 to 6. We found a clear trend in most cases that growing a large forest is much more important than tall trees: indeed, in many cases 1000 tree forests with 1 level perform equally well to those with 6 levels. Figure \ref{fig:CRFperformance} in Appendix \ref{sec:CRFperformance} plots the results. The depth of trees only appears relevant in large forests for the \texttt{iono}, \texttt{magic} and two \texttt{musk} data sets.  This may indicate the data sets with more non-linearities, a hypothesis supported by the poor performance of the linear methods in these cases.  Indeed, 1 level deep trees appear to be good candidates for additive models.

\subsubsection{Stochastic fraction}
\label{sec:RFfracPerf}

The CRF algorithm presented in \S\ref{sec:RF} has the option, presented in \S\ref{sec:RFfrac}, of including an unbiased encrypted stochastic fraction estimate in order to reduce the amount of shrinkage that small leaves undergo.  To analyse the impact that this has on the performance, the AUC was recomputed after setting different values for $M$ from 0 (i.e.~original algorithm, no stochastic fraction) through to 64 in all 100 train/test splits of the 20 datasets that were presented already.

\begin{figure}[!ht]
\captionsetup[subfigure]{belowskip=-1cm}
\centering
\includegraphics[width=\textwidth]{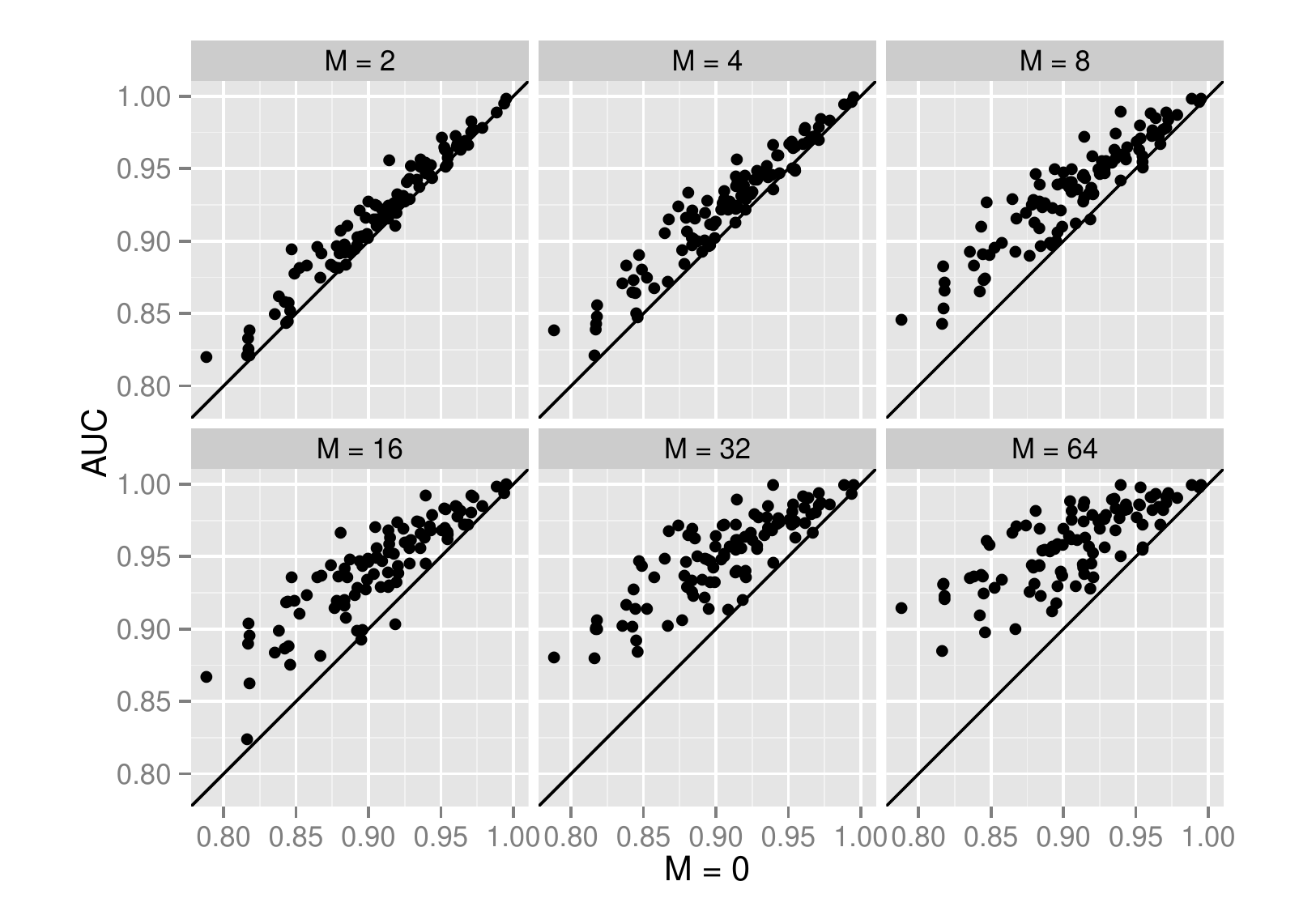}
\vspace{-0.5cm}
\caption{The AUC change for \texttt{iono} data set with different values for $M$ in the stochastic fraction estimate.  $x$-axis is always with no stochastic fraction estimate; $y$-axis is for shown value of $M$; one point per train/test split.
}
\label{fig:ionoSF}
\end{figure}

Figure \ref{fig:ionoSF} shows the results from the data set with the largest improvement in AUC performance from among the 20 data sets when using the stochastic fraction estimate.  Improvements in AUC of up to 12.6\% were achieved using the stochastic fraction versus omitting it.  All points which are above the $y=x$ line indicate improvements in AUC when using the stochastic fraction for the particular train/test split.

\begin{figure}[!ht]
\captionsetup[subfigure]{belowskip=-1cm}
\centering
\includegraphics[width=\textwidth]{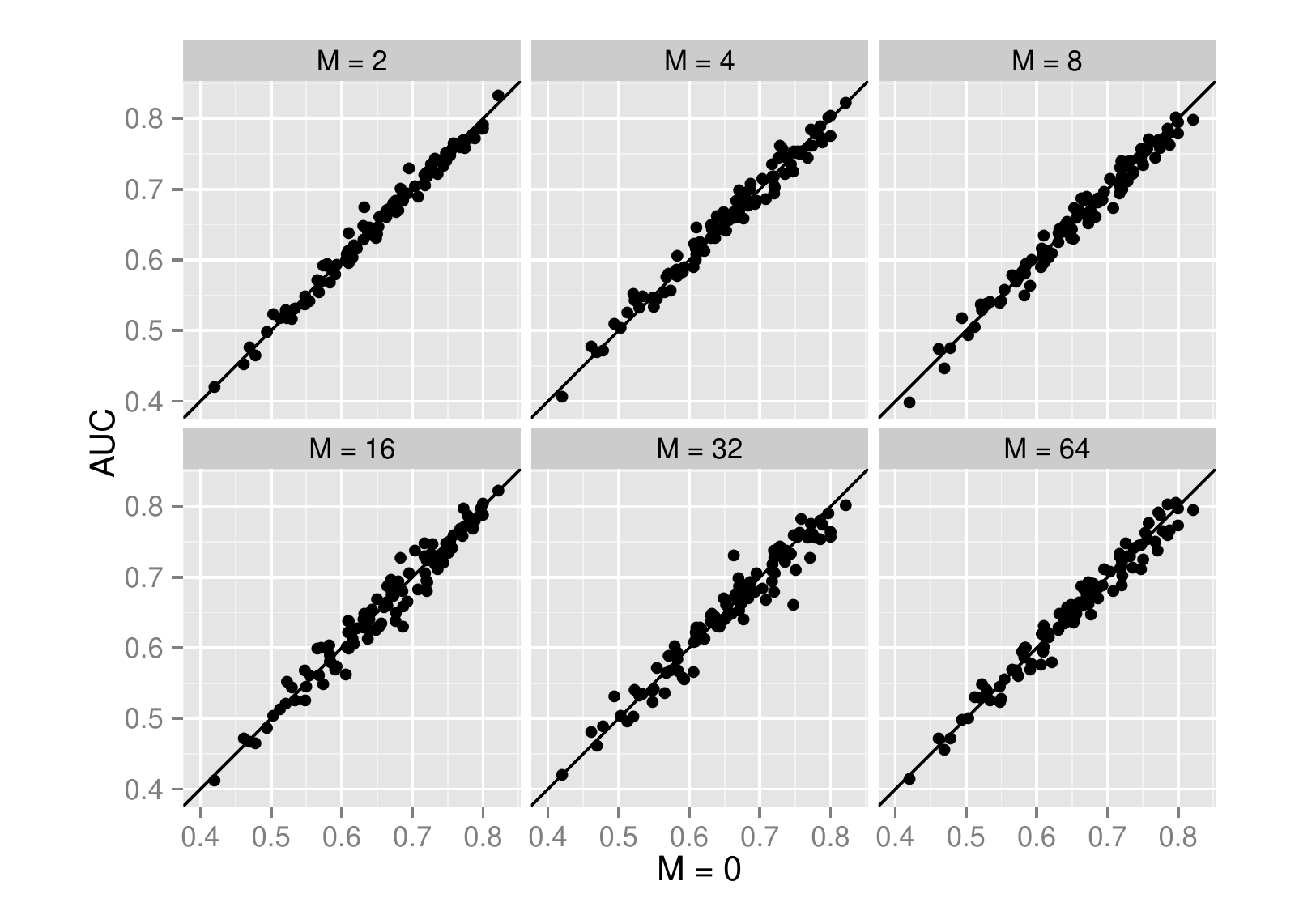}
\vspace{-0.5cm}
\caption{The AUC change for \texttt{bcw\_p} data set with different values for $M$ in the stochastic fraction estimate.  $x$-axis is always with no stochastic fraction estimate; $y$-axis is for shown value of $M$; one point per train/test split.
}
\label{fig:bcw_pSF}
\end{figure}

None of the 20 tested datasets decreased in average AUC with increasing $M$, but Figure \ref{fig:bcw_pSF} shows the data set for which the AUC was least improved.  In this instance it is striking that all the points cluster around the $y=x$ line, showing that in essence in the worst case the stochastic fraction estimate has essentially negligible impact.

These figures empirically illustrate the fact that the stochastic fraction has the potential to dramatically improve the performance of completely random forests, whilst not really having a negative impact in those situations where it does not help.  As such, it would seem to make sense to include by default.

\subsection{Case study in encrypted cloud computing machine learning}
\label{sec:AWS}

To demonstrate the potential to utilise cloud computing resources for sensitive data analysis we undertook a benchmark case study performed with fully encrypted data analysis on the original Wisconsin breast cancer data set using a compute cluster of 1152 CPU cores on Amazon Web Services, at a total cost of less than US\$\,24 at the time of writing.  The resources used here are readily available to any scientist.

\subsubsection{The problem setup}

As mentioned earlier in \S\ref{sec:CRFparallel}, the completely random forest is amenable to embarrassingly parallel computation, whereby the data can be split into shards using the same random seed, the expensive fitting step performed on each shard, and the final fit produced by inexpensively summing the individual tree shards.

The training part of the Wisconsin data ($n=547$) were initially split into shards of at most 32 observations each, resulting in 17 full 32-observation shards and an 18th shard of 3 observations.  These shards were then each encrypted using the \texttt{HomomorphicEncryption} R package \citep{HEpkg} under the Fan and Vercauteren scheme using parameters:
\begin{align*}
  \Phi &= x^{8192}+1 \\
  q &= 2^{224} \\
    &= 26959946667150639794667015087019630673637144422540572481103610249216 \\
  t &= 200000 \\
  \sigma &= 16
\end{align*}
This renders a theoretical cipher text size of $2 \times 8192 \times 224 \div (8 \times 1024) = 448$KB per encrypted value, ignoring keys and overhead.  These parameters offer around 158-bits of security (using bounds in \citet{lindner2011better}) --- informally this means on the order of $2^{158}$ fundamental operations must be performed on average in order to break the encryption.  On disk, each gzipped shard of 32 cipher texts for the predictors occupied about 737MB and for the responses occupied about 33.7MB, for a total disk space of around 13.8GB.

This data was uploaded to an Amazon S3 bucket, with the transfer time using the University internet connection being approximately 16 minutes.  If this data was to be stored long-term on Amazon S3, it would cost US\$\,0.42 per month at the time of writing.

Once the data was in place, an Amazon SQS queue was setup in which to store a reference to each shard.  This queue acts as a simple job dispatch system, designed to ensure that each server in the cluster can remain completely independent for maximum speed, by eliminating inter-server communication and as a mechanism to ensure no duplication of work.

With these elements in place, the RStudio AMI \texttt{ami-628c8a0a} \citep{ami} was extended to add a startup script which (in summary) fetches the work to perform from the SQS queue, downloads it from S3, executes the forest building using the \texttt{EncryptedStats} R package, and uploads the result to the S3 bucket.

\subsubsection{The fitting run}

The fitting run used Amazon's spot instances: these are a `stock market' for unused capacity, where it is often possible to bid below the list price for compute servers.  The completely random forest is well suited to exploiting low spot prices on EC2 wherever they may arise because it can be formulated in an embarrassingly parallel manner and launched in very geographically dispersed regions without regard for connectivity speeds, since communication costs between nodes are effectively zero.

When the run was performed on 5$^\mathrm{th}$ May 2015, the spot price for c3.8xlarge instances was lowest in Dublin, Ireland and S\~{a}o Paulo, Brazil.  Consequently, the data was replicated to two S3 buckets local to these regions and the customised AMI copied.  Then a cluster of 18 c3.8xlarge servers was launched in each region, giving a total of $1,152$ CPU cores and $2,160$GB of RAM.

Each server was setup to compute 50 trees on its shard of data and every shard was handed out twice so that a total of 100 trees were fitted.  Tree growth for the two different sets of 50 trees were initialised from a common random seed, eliminating the need for servers to communicate the trees grown.

After 1 hour and 36 minutes the cluster had completed the full run and finished uploading the encrypted tree fit (that is, encrypted versions of $\bar{\rho}^t_{bc}\ \forall\,t,b,c$) back to the S3 bucket.  The total space required for storing the 36 forests of 50 trees fitted on each shard was 15.6GB.  At this juncture, the forests could be combined homomorphically to produce a single forest of 100 trees which would then require 868MB to store.  Note that with the tree fitted, it would then be possible to archive the 13.8GB of original data so that only 868MB needs to be stored long term or downloaded.

The cost of the 36 machines for 2 hours was US\$\,23.86 (about \pounds\,15.66).  Note that the spot prices were not exceptionally low on the day in question and no effort was made to select an opportune moment.  By the same token the price is inherently variable and it may be necessary to wait a short time for favourable spot prices to arise if there are none at the time of analysis.

\subsubsection{Results}

The encrypted version of the forest was downloaded, decrypted and compared to the results achieved when performing the same fit using an unencrypted version of the data, starting tree growth from the same seeds and using identical R code from the \texttt{EncryptedStats} package (separate code is not required due to the \texttt{HomomorphicEncryption} package fully supporting operator overloading).  The resultant fit from both encrypted and unencrypted computation was in exact agreement.

\section{Discussion}
\label{sec:discussion}

Fully homomorphic encryption schemes open up the prospect of privacy preserving machine learning applications. However practical constraints of existing FHE schemes demand tailored approaches. With this aim we made bespoke adjustments to two popular machine learning methods, namely extremely random forests and naive Bayes classifiers, and demonstrated their performance on a variety of classifier learning tasks. We found the new methods to be competitive against their unencrypted counterparts. 

To the best of our knowledge these represent the first machine learning schemes tailored explicitly for homomorphic encryption so that all stages (fitting and prediction) can be performed encrypted without any multi-party computation or communication.

Furthermore, the \emph{unencrypted} version of these new methods will scale to massive data sets, because the most complex operations involved are addition and multiplication.  Indeed, even these simple operations can be performed in parallel for most of the presented algorithms and map directly to CPU vector instructions.  This is an interesting avenue of future research.

\section*{Acknowledgements}

The authors would like to thank the EPSRC and LSI-DTC for support.  Louis Aslett and Chris Holmes were supported by the i-like project (EPSRC grant reference number EP/K014463/1).  Pedro Esperan\c{c}a was supported by the Life Sciences Interface Doctoral Training Centre doctoral studentship (EPSRC grant reference number EP/F500394/1).

\renewcommand\bibname{References}
\bibliography{FHE_paper}
\bibliographystyle{Louis_agsm}

\appendix
\section{Completely Random Forest (CRF) algorithm}
\label{sec:CRFalg}

In detail, consider a training set with $N$ observations consisting of a categorical response, $y_i \in \mathcal{C}$, and $P$ predictors, $x_{ip}$, (categorical / ordinal / continuous), for $i \in \{1, \dots, N\}, p \in \{1, \dots, P\}$.  All variables (predictors and response) are first transformed using method 1 from \S\ref{sec:Quantisation} prior to encryption.  Thus, $y_i \to \tilde{y}_{ic} \in \{0,1\}$ for $c \in \{ 1, \dots, |\mathcal{C}| \}$ and $x_{ip} \to \tilde{x}_{ipk}$.  Consider these herein to be in encrypted form.

\begin{figure}
\makebox[\linewidth]{
  \includegraphics[keepaspectratio=true,angle=90, scale=0.875]{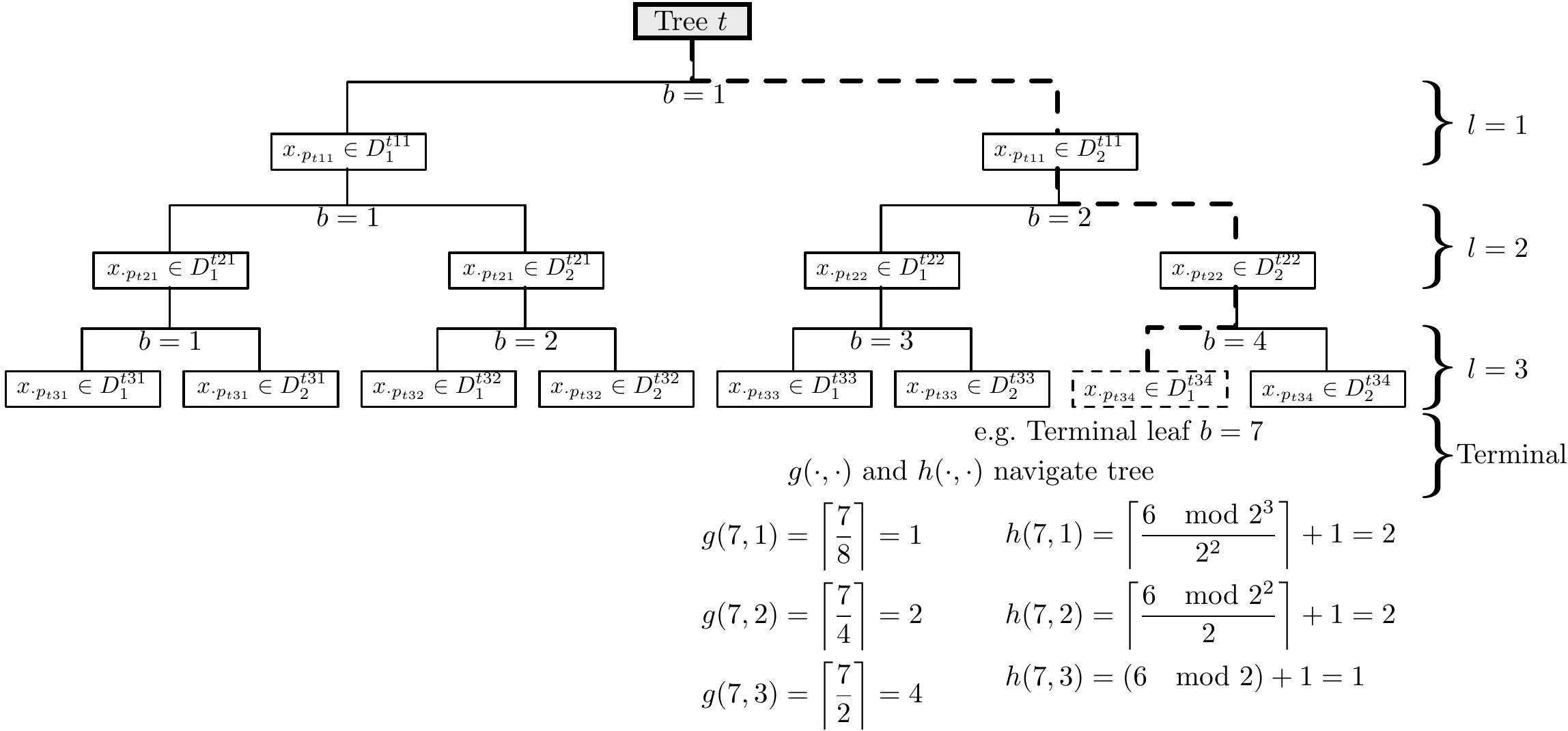}}
\captionof{figure}{Completely Random Forest algorithm depiction.  The dashed line shows the evaluation of a training/testing observation via the $g$ and $h$ functions.  $b$ signifies the branch number at the given level $l$.}\label{fig:SF}
\end{figure}

The proposed algorithm for Completely Random Forests (CRFs) is then as follows:

\noindent\textbf{Algorithm}
\begin{enumerate}
  \item Specify the number of trees to grow, $T$, and the depth to make each tree, $L$.
  \item For each $t \in \{1, \dots, T\}$, build a tree in the forest:
    \begin{enumerate}
      \item \textbf{Tree growth:} For each $l \in \{1, \dots, L\}$, build a level:
        \begin{enumerate}
          \item Level $l$ will have $2^{l-1}$ branches (splits), each of which will have a partition applied as follows.  For each $b \in \{1, \dots, 2^{l-1}\}$, construct the partitions:
            \begin{enumerate}
              \item \textbf{Splitting variable:} Select a variable $p_{tlb}$ at random from among the $P$ predictors.  Due to the encoding of \S\ref{sec:Quantisation}, this variable has a partition $\mathcal{K}_{p_{tlb}}$ associated with it.
              \item \textbf{Split point:} Create a partition of $\mathcal{K}_{p_{tlb}}$ at random in order to perform a split on variable $p_{tlb}$, $\mathcal{D}_{tlb} = \{D^{tlb}_1, D^{tlb}_2\}$ where each $D^{tlb}_j = \bigcup K^{p_{tlb}}_i$ for some $K^{p_{tlb}}_i \in \mathcal{K}_{p_{tlb}}$, with $D^{tlb}_1 \cap D^{tlb}_2 = \varnothing$ and $D^{tlb}_1 \cup D^{tlb}_2 = \bigcup_{\forall\,i} K^{p_{tlb}}_i$.  Note that for categorical predictors this is a random assignment of levels from the partition $\mathcal{K}_{p_{tlb}}$ to each $D^{tlb}_j$, while for ordinal predictors a split point is chosen and the partition formed by the levels either side of the split.
              
              Note also the indexing of $\mathcal{D}_{tlb}$ to emphasise if variable $p_{tlb}$ is selected more than once (in different levels or trees) a different random split is chosen.
            \end{enumerate}
        \end{enumerate}

      \item \textbf{Tree fitting:} The total number of training observations belonging to category $c$ in the completely randomly grown tree $t$ at terminal leaf $b\in\{1,\dots,2^L\}$ is then:
        \[ \bar{\rho}^{t}_{bc} = \sum_{i=1}^N \tilde{y}_{ic} \prod_{l=1}^L \left( \sum_{k \in D^{tlg(b,l)}_{h(b,l)}} \tilde{x}_{i,\,p_{tlg(b,l)},\,k} \right) \]
        where
        \begin{align*}
          g(b,l) &:= \left\lceil\frac{b}{2^{L+1-l}} \right\rceil \\
          h(b,l) &:= \left\lfloor\frac{(b-1) \mod 2^{L+1-l}}{2^{L-l}} \right\rfloor + 1 
        \end{align*}
        
        Figure \ref{fig:SF} (page \pageref{fig:SF}) is useful for understanding this.
        
        Note in particular that written this way $\bar{\rho}^{t}_{bc}$ is simply a polynomial and can be computed homomorphically with multiplicative depth $L$.  Both $g(\cdot, \cdot)$ and $h(\cdot, \cdot)$ involve only indices of the algorithm which will not be encrypted.  Thus, the training data can be evaluated on the tree without the use of comparisons.
        
        The total number of training observations in this terminal leaf is then simply:
        \[ \underline{\rho}^t_b = \sum_{c=1}^{|\mathcal{C}|} \bar{\rho}^t_{bc} \]
        
        Thus a single tree, $t$, fitted to a set of training data, is defined by the tuple of sets:
        \begin{align*}
          &\{ p_{tlb} : l=\{1,\dots,L\}, b=\{1,\dots,2^{l-1} \}\} \\
          &\{ \mathcal{D}_{tlb} : l=\{1,\dots,L\}, b=\{1,\dots,2^{l-1}\}\} \\
          &\{\bar{\rho}_{bc}^t : b=\{1,\dots,2^{L-1}\}, c=\{1,\dots,|\mathcal{C}|\}\}
        \end{align*}
    \end{enumerate}

  \item \textbf{Prediction:} Once the forest has been grown, attention turns to prediction.  Given an encrypted test observation with predictors $\tilde{x}^\star_{jk}$, the objective is to predict the response category. 
    Define $\hat{y}^\star_c$ to be the number of votes for response category $c$, which can be simply computed as:
    \[ \hat{y}^\star_c = \sum_{t=1}^T \sum_{b=1}^{2^L} \bar{\rho}_{bc}^t \prod_{l=1}^L \left( \sum_{k \in D^{tlg(b,l)}_{h(b,l)}} \tilde{x}_{p_{tlg(b,l)},\,k}^\star \right) \]
    where $g(\cdot, \cdot)$ and $h(\cdot, \cdot)$ are as defined above.  Each $\hat{y}^\star_c$ is returned from the cloud to the client.  The client decrypts and forms a predictive empirical `probability' as:
    \begin{equation} \hat{p}^\star_c = \frac{\hat{y}^\star_c}{\sum_{c=1}^{|\mathcal{C}|} \hat{y}^\star_c} \label{eq:SFpredprob2}\end{equation}
\end{enumerate}

Hence, the proposed CRF algorithm makes use of a quantisation procedure on the data, followed by completely random selection of variable and completely random partition on the quantile bins, in order to eliminate any need to perform comparisons.

Tree growth (2a) occurs unencrypted and is a very fast operation.

Tree fitting (2b) involves counting the number of training observations which lie in each terminal node of each tree.  In computing $\bar{\rho}^{t}_{bc}$, the inner sum evaluates whether an observation has predictor value in the relevant partition at level $l$ of tree $t$ (by \eqref{eq:range}).  The product over all levels then results in 1 if and only if the observation lies in this path through the tree.  Finally, multiplication by $\tilde{y}_{ic}$ ensures the observation only counts if it belongs to response category $c$.

If growing a tree encrypted is the objective then at this point the resulting $\bar{\rho}^{t}_{bc}$ can be decrypted.  Alternatively, without decrypting it is possible to immediately predict with the tree, using a similar procedure of evaluating an effectively binary circuit representation of the tree and data.

\newpage

\section{Theoretical parameter requirements for \citet{Fan12}}
\label{sec:pars}

\subsection{Completely Random Forests parameter requirements}

In order to be able to fit extremely random forests of $T$ trees each of depth $L$ encrypted under the scheme of \citet{Fan12}, the parameter choice must support at least $L$-depth multiplications.  Additionally, the message space must support coefficient values up to size $T \max \{ \sum_i \tilde{y}_{ic} : c=1, \dots, |\mathcal{C}| \}$, since all values will accumulate on the first coefficient of $\mathring{m}(x)$.  This covers the most extreme (and improbable) possibility that a leaf in every tree contains all the observations for a particular class.

In the event that the stochastic fraction variant is used, the parameters must support a total of at least $L+M$ multiplications. If prediction is to be performed without decrypting the tree or refreshing the noise in the tree cipher text, then the scheme must support an additional $L$ multiplications on top of the values above.

The function \texttt{parsHelp} in the \texttt{HomomorphicEncryption} package \citep{HEpkg} can be used to guide selection of the scheme parameters $d, t, q,$ and $\sigma$ which will enable the multiplicative depth and maximum coefficient values required.

\subsection{Semiparametric Na\"{i}ve Bayes parameter requirements}

In order to be able to fit SNB encrypted under the scheme of \citet{Fan12}  \citep*{Part1}, the coefficients $\{ e_{j} , d_{j} \}_{j \in \mathbb{N}_{1:P}}$ must be returned for offline assembly via Equation~\eqref{eq:SNB_predict_b} (because homomorphic divisions are not possible), together with the empirical class frequencies. 
The former have multiplicative depth equal to 4 and 3, respectively, and the later equal to 0.
With respect to the growth in message coefficient size for the \texttt{FandV} scheme, the biggest growth comes from computing $d_{j}$ in Equation~\eqref{eq:1step_d}.  As described in detail in \citet*{Part1}, initially the polynomial representation of all $x_{ij}$ will have maximum coefficient value of 1.  Therefore, by the multinomial theorem the maximum coefficient value of the polynomial representation of $x_{ij}^2$ will be 2.  Thus the maximum coefficient for the first term is $2N$.  In contrast, the maximum coefficient of $\sum_{i} x_{ij}$ will be $N$, so that by the multinomial theorem the maximum coefficient value of the polynomial representation of $\left( \sum_{i} x_{ij} \right)^2$ will be $2N^2$.  Therefore to ensure correct decryption after fitting the naïve Bayes model the message space must support coefficient values up to size $2N^2$.

The exclusion of the intercept term has an impact on these requirements and, consequently, on the running time of the algorithm. 
Precisely, numerator and denominator of $\hat{\beta}$ have multiplicative depth equal to 2 and 1, respectively, in this case. The message coefficient requirements will also be significantly relaxed, with the maximum coefficient value being $2N$.

\section{Multinomial Na\"{i}ve Bayes algorithm}
\label{sec:MNB}

The Multinomial Na\"{i}ve Bayes (MNB) classifier estimates $\mathbb{P}(x_{j}|y)$ and $\mathbb{P}(y)$ using the corresponding empirical probabilities,
\begin{equation}
    \hat{p}_{x_{j}=k|y=c} = \frac{\#\{X_{\smallbullet j}=k \wedge Y=c\}} {\#\{Y=c\}}
\qquad\text{and}\qquad 
\hat{p}_{y=c} = \frac{\#\{Y=c\}} {N}
\end{equation}
where $\#\{z\}$ is the cardinality of the set satisfying condition $z$; 
$k$ denotes the $k$th possible value taken by $x_{ij} \in \mathbb{N}_{1:M_{j}}$;
and $X_{\smallbullet j}$ and $Y$ are $N$-dimensional vectors of observations for predictor $j$ and response, respectively.
The ``Multinomial'' qualifier should be understood to mean precisely that $x_{ij}$ takes values in the discrete set $\mathbb{N}_{1:M_{j}}$ and so \mbox{$\mathcal{X} = \mathbb{N}_{1:M_{1}} \times \dots \times \mathbb{N}_{1:M_{P}}$}.

Because it is not possible to perform homomorphic divisions, the algorithms will be written in terms of unnormalised probabilities, or counts:
\begin{equation}
    \hat{p}_{x_{j}=k|y=c}^{(u)} = \#\{X_{\smallbullet j}=k \wedge Y=c\}
\qquad\text{and}\qquad 
\hat{p}_{y=c}^{(u)} = \#\{Y=c\}.
\end{equation}
Estimation, therefore, requires the computation of $\hat{p}_{x_{j}|y}^{(u)}$ and $\hat{p}_{y}^{(u)}$ for all possible values of $y$  and $x_{j}$, and for all $j \in \mathbb{N}_{1:P}$.
This corresponds to a $2 \times M_{j}$ table of conditional probabilities for each predictor $j \in \mathbb{N}_{1:P}$; plus two class probabilities. In total, this makes $2 + 2 \cdot \sum_{j=1}^{P} M_{j}$ parameters. 

A computational problem arises from the fact that $\hat{p}_{x_{j}|y}^{(u)}$ can not be computed directly: since the values $x_{ij}$ are encrypted, this makes it impossible to count, say, the number of 2s in the vector $X_{\smallbullet j}$ directly. 
An indirect route follows from the first quantisation method in Section~\ref{sec:Quantisation}. 
Let $\tilde{X}$ denote the binary-expanded version of $X$;
and $\tilde{x}_{ijk}$ denote the $k$th bin of $\tilde{X}_{ij\smallbullet} = ( \tilde{x}_{ij1}, \dots, \tilde{x}_{ijM_{j}} )$. 
Then, the conditional counts can be computed as
\begin{align}\label{eq:MNB_countsX}
    \hat{p}_{x_{j}=k|y=1}^{(u)} = \sum_{i=1}^{N} \tilde{x}_{ijk} y_{i} \qquad\text{and}\qquad
    \hat{p}_{x_{j}=k|y=0}^{(u)} = \sum_{i=1}^{N} \tilde{x}_{ijk} (1-y_{i}).
\end{align}
and the class counts as
\begin{align}
    \hat{p}_{y=1}^{(u)} = \sum_{i=1}^{N} y_{i} \qquad\text{and}\qquad
    \hat{p}_{y=0}^{(u)} = N - \sum_{i=1}^{N} y_{i}.
\end{align}
The multiplicative depth of $\hat{p}_{y}^{(u)}$ is $0$, while that of $\hat{p}_{x_{j}|y}^{(u)}$ is $1$.  As standard practice, implementations should use Laplace-smoothed counts or probabilities to avoid the problems arising from having zero counts for some values of $k$.

Classification of an observation $X^{\star} = \{x_{j}^{\star}\}_{j \in \mathbb{N}_{1:P}}$ requires the computation of the factors
\begin{equation}\begin{aligned}
    F_{c} 
    &= \mathbb{P}(y=c) \prod_{j=1}^P \mathbb{P}(x_{j}^{\star}|y=c)  \\
    &= N^{-1} 
    \left(\hat{p}_{y=c}^{(u)} \right)^{-(P-1)} 
    \prod_{j=1}^P \hat{p}_{x_{j}^{\star}|y=c}^{(u)}
\end{aligned}\end{equation}
for $c \in \{0,1\}$.
If $X^{\star}$ is known only in encrypted format, then $\hat{p}_{x_{j}^{\star}|y=c}^{(u)}$ must be computed indirectly as
\begin{equation}\label{eq:MNB_countsXstar}
    \hat{p}_{x_{j}^{\star}|y=c}^{(u)} = \sum_{k=1}^{M_{j}} \tilde{x}_{jk}^{\star} \cdot \hat{p}_{x_{j}=k|y=c}^{(u)}
\end{equation}
where $\tilde{X}_{j}^{\star} = \{ \tilde{x}_{jk}^{\star} \}_{k \in \mathbb{N}_{1:M_{j}}}$ is the binary expanded version of $x_{j}^{\star}$; 
and which has multiplicative depth equal to $2$.
The predicted probability of $X^{\star}$ belonging to class $y=1$ is then 
\mbox{$\mathbb{P}(y=1|X^{\star}) = F_{1} / (F_{0}+F_{1})$}, a final step which must be computed offline.

It should be noted that homomorphic MNB requires the use of binary-expanded data ($\tilde{X}$) as opposed to integer data ($X$).
This has a direct impact on the computational cost (as it leads to a larger number of homomorphic operations; recall Equations \eqref{eq:MNB_countsX} and \eqref{eq:MNB_countsXstar}) and memory requirements (as the data must be represented in the binary-expanded form).

In the next Section (\S\ref{sec:NB_SNB}) an alternative approach is suggested which mitigates some of these problems, while still relying on NB assumption.

\subsection*{Theoretical parameter requirements for \citet{Fan12}}

For MNB, the elements to be returned are
$\hat{p}_{y=c}^{(u)}$
and
$ \hat{p}_{x_{j}^{\star}|y=c}^{(u)}$, 
for $j \in \mathbb{N}_{1:P}$ and $c \in \{0,1\}$, with the former having multiplicative depth equal to $0$ and the later equal to $2$. 

It is possible reduce the number of elements to be returned to 4 by computing 
$(\hat{p}_{y=c}^{(u)})^{P-1}$ 
and 
$\prod_{j=1}^{P} \hat{p}_{x_{j}^{\star}|y=c}^{(u)}$ 
homomorphically but this comes at the cost of an increase in the multiplicative depth to $P-1$ and $3P-1$, respectively.

\section{Open source software contributions}
\label{sec:APDX_implementation}

We have contributed two open source R packages which make working with the techniques described and contributed herein easy.

The first, \texttt{HomomorphicEncryption} \citep{HEpkg}, implements the \citet{Fan12} homomorphic encryption scheme in a generic framework which can be extended to plug in other encryption schemes over time.  The package is written in high performance C++ and utilises full operator overloading in R.  Indeed, all common types such as vectors and matrices are implemented so that they can be transparently encrypted and manipulated just like their unencrypted counterparts.  The theoretical security and multiplicative depth bound results of different papers are wrapped in a helper function to aid in parameter choice.

The second, \texttt{EncryptedStats} \citep{ESpkg}, implements the tailored machine learning methods of this paper in such as way as to enable fitting and prediction on unencrypted and encrypted data using precisely the same code paths (thanks to the operator overloading support of \texttt{HomomorphicEncryption}).

\subsection{Introduction to the \texttt{HomomorphicEncryption} package}
\label{App:HEpkg}

There are three steps to encrypting data using the package: i) parameter selection; ii) key generation; and iii) encryption/decryption.  Once these are understood, then homomorphic operations follow naturally from operator overloading.

\subsubsection*{Parameter selection}

The scheme of \citet{Fan12} has four parameters which must be selected.  In the package these are referred to as:
\begin{itemize}
  \item \texttt{d}, the power of the cyclotomic polynomial ring (default 4096);
  \item \texttt{sigma}, the standard deviation of the discrete Gaussian used to induce a distribution on the cyclotomic polynomial ring (default 16.0);
  \item \texttt{qpow}, the power of 2 to use for the coefficient modulus (default 128);
  \item \texttt{t}, the value to use for the message space modulus (default 32768).
\end{itemize}
which are arguments of the \texttt{pars} function, the first argument of which selects the encryption scheme to use.  Presently only \citet{Fan12} is implemented (\texttt{"FandV"}).  Say the defaults are acceptable except for the message space modulus which should be doubled, it could be set by running:
\begin{Shaded}
\begin{Highlighting}[]
\NormalTok{> p <-}\StringTok{ }\KeywordTok{pars}\NormalTok{(}\StringTok{"FandV"}\NormalTok{, }\DataTypeTok{t=}\DecValTok{65536}\NormalTok{)}
\NormalTok{> p}
\NormalTok{Fan and Vercauteren parameters}
\NormalTok{phi = x^4096+1}
\NormalTok{q = 340282366920938463463374607431768211456 (128-bit integer)}
\NormalTok{t = 65536}
\NormalTok{delta = 5192296858534827628530496329220096}
\NormalTok{sigma = 16}
\NormalTok{Security level approx 128-bits}
\NormalTok{Supports multiplicative depth of 3 with overwhelming probability}
\NormalTok{(i.e. lower bound, likely more possible)}
\end{Highlighting}
\end{Shaded}

The selection of these parameters can be bewildering, so a feature of the package is a parameter helper function \texttt{parsHelp} which has a simpler set of options and will use theoretical bounds from the cryptography literature to automatically select the parameters for you.  These simpler choices are:
\begin{itemize}
  \item \texttt{lambda}, the security level required (bits), default is 80;
  \item \texttt{max}, the largest absolute value you will need to store encrypted, default is 1000;
  \item \texttt{L}, the deepest multiplication depth you need to be able to evaluate encrypted, default is 4.
\end{itemize}
So, for example, we could have selected the ability to store a larger message, but requiring at least 140 bits of security:
\begin{Shaded}
\begin{Highlighting}[]
\NormalTok{> p <-}\StringTok{ }\KeywordTok{parsHelp}\NormalTok{(}\StringTok{"FandV"}\NormalTok{, }\DataTypeTok{lambda=}\DecValTok{140}\NormalTok{, }\DataTypeTok{max=}\DecValTok{65536}\NormalTok{)}
\NormalTok{> p}
\NormalTok{Fan and Vercauteren parameters}
\NormalTok{phi = x^8192+1}
\NormalTok{q = 365375409332725729550921208179070754913983135744 (158-bit integer)}
\NormalTok{t = 131072}
\NormalTok{delta = 2787593149816327892691964784081045188247552}
\NormalTok{sigma = 16}
\NormalTok{Security level approx 273-bits}
\NormalTok{Supports multiplicative depth of 4 with overwhelming probability}
\NormalTok{(i.e. lower bound, likely more possible)}
\end{Highlighting}
\end{Shaded}

Note that the parameter helper will often be conservative in setting the security level: the theoretical bounds may not allow all constraints to be simultaneously satisfied exactly and so the security is treated as a minimum required level.

\subsubsection*{Key generation}

With the parameter values stored in the output from either the \texttt{pars} or \texttt{parsHelp} functions, generating the keys is then straight-forward:
\begin{Shaded}
\begin{Highlighting}[]
\NormalTok{> keys <-}\StringTok{ }\KeywordTok{keygen}\NormalTok{(p)}
\end{Highlighting}
\end{Shaded}
The public key is stored in \texttt{keys\$pk}, the secret key in \texttt{keys\$sk} and the relinearisation key (used during multiplication operations) in \texttt{keys\$rlk}.

In order to save the keys for future use (or to give the public key to another party), see the \texttt{saveFHE} and \texttt{loadFHE} functions in the package.

\subsubsection*{Encryption/Decryption}

Encryption and decryption take place according to the same format as in \S\ref{sec:HE}, with encryption requiring only the public key and decryption only the secret key:
\begin{Shaded}
\begin{Highlighting}[]
\NormalTok{> ct <-}\StringTok{ }\KeywordTok{enc}\NormalTok{(keys$pk, }\DecValTok{2}\NormalTok{)}
\NormalTok{> ct}
\NormalTok{Fan and Vercauteren cipher text}
\NormalTok{( c_0 = 13x^4096+137352383756050088155497465542996641770x^4095+7041424...,}
\NormalTok{c_1 = -46602244866771058520503760576262076531x^4095-4071136373586351... )}

\NormalTok{> m <-}\StringTok{ }\KeywordTok{dec}\NormalTok{(keys$sk, ct)}
\NormalTok{> m}
\NormalTok{[1] 2}
\end{Highlighting}
\end{Shaded}

\subsubsection*{Homomorphic operations}

Using homomorphic operations is the simplest aspect of all, because the package operator overloads $+$ and $\times$ (\texttt{*}) operations and also implements native support for vectors and matrices being encrypted to cipher texts.

The following example mixes scalars, vectors and matrices and showcases both scalar and matrix operations:
\begin{Shaded}
\begin{Highlighting}[]
\NormalTok{> x <-}\StringTok{ }\DecValTok{2}
\NormalTok{> y <-}\StringTok{ }\KeywordTok{c}\NormalTok{(}\DecValTok{1}\NormalTok{,}\DecValTok{2}\NormalTok{,}\DecValTok{3}\NormalTok{)}
\NormalTok{> z <-}\StringTok{ }\KeywordTok{matrix}\NormalTok{(}\DecValTok{1}\NormalTok{:}\DecValTok{9}\NormalTok{, }\DataTypeTok{nrow=}\DecValTok{3}\NormalTok{)}
\NormalTok{> z}
\NormalTok{     [,1] [,2] [,3]}
\NormalTok{[1,]    1    4    7}
\NormalTok{[2,]    2    5    8}
\NormalTok{[3,]    3    6    9}
\NormalTok{> xct <-}\StringTok{ }\KeywordTok{enc}\NormalTok{(keys$pk, x)}
\NormalTok{> yct <-}\StringTok{ }\KeywordTok{enc}\NormalTok{(keys$pk, y)}
\NormalTok{> zct <-}\StringTok{ }\KeywordTok{enc}\NormalTok{(keys$pk, z)}
\NormalTok{> res <-}\StringTok{ }\NormalTok{x +}\StringTok{ }\NormalTok{y %*%}\StringTok{ }\KeywordTok{t}\NormalTok{(z) %*%}\StringTok{ }\NormalTok{z}
\NormalTok{> resct <-}\StringTok{ }\NormalTok{xct +}\StringTok{ }\NormalTok{yct %*%}\StringTok{ }\KeywordTok{t}\NormalTok{(zct) %*%}\StringTok{ }\NormalTok{zct}
\NormalTok{> res}
\NormalTok{     [,1] [,2] [,3]}
\NormalTok{[1,]  230  554  878}
\KeywordTok{> dec}\NormalTok{(keys$sk, resct)}
\NormalTok{     [,1] [,2] [,3]}
\NormalTok{[1,]  230  554  878}
\end{Highlighting}
\end{Shaded}

\subsection{Introduction to the \texttt{EncryptedStats} package}

The first two methods implemented in the package are the two methods presented in this paper.

\subsubsection*{Completely Random Forests}

There are three steps in working with CRFs in the package: i) grow a forest; ii) fit data to each tree (accumulate votes); and iii) predict a new observation.

When working with CRFs, the data must already be in the representation of Method 1 from \S\ref{sec:Quantisation} in either an encrypted or unencrypted matrix in R.

Imagine a toy data set consisting of $N$ observations of two variables (\texttt{v1} and \texttt{v2}) where the first is ordinal and the second categorical.  Assume that these are in quintiles so that the data matrix is $N \times 10$, stored in the variable called \texttt{X}, say, in R.  Also assume the responses of the training observations are similarly stored in \texttt{y}.

\vspace{6pt}\noindent\textbf{i) Forest growth}

Tree growth is blind from the data, so you only need to identify the variable names of each column of the (un)encrypted data matrix and their type --- either ordinal (\texttt{"ord"}) or categorical (\texttt{"cat"}).  For the toy example,
\begin{Shaded}
\begin{Highlighting}[]
\NormalTok{> Xvars <-}\StringTok{ }\KeywordTok{rep}\NormalTok{(}\KeywordTok{c}\NormalTok{(}\StringTok{"v1"}\NormalTok{, }\StringTok{"v2"}\NormalTok{), }\DataTypeTok{each=}\DecValTok{5}\NormalTok{)}
\NormalTok{> Xvars}
\NormalTok{[1] "v1" "v1" "v1" "v1" "v1" "v2" "v2" "v2" "v2" "v2"}
\end{Highlighting}
\end{Shaded}
This vector of names is used to tell the tree growth algorithm that the first 5 columns of \texttt{X} contain a quantisation of variable \texttt{v1} and the second 5 columns contain a quantisation of variable \texttt{v2}.
\begin{Shaded}
\begin{Highlighting}[]
\NormalTok{> Xtype <-}\StringTok{ }\KeywordTok{c}\NormalTok{(}\StringTok{"v1"}\NormalTok{=}\StringTok{"ord"}\NormalTok{, }\StringTok{"v2"}\NormalTok{=}\StringTok{"cat"}\NormalTok{)}
\NormalTok{> Xtype}
\NormalTok{   v1    v2 }
\NormalTok{"ord" "cat" }
\end{Highlighting}
\end{Shaded}
This vector then specifies the type for each variable.  Together these can be passed to the forest growth algorithm, along with the specification of the number of trees to grow (\texttt{T}) and how deep to grow the trees (\texttt{L}):
\begin{Shaded}
\begin{Highlighting}[]
\NormalTok{> forest <-}\StringTok{ }\KeywordTok{CRF.grow}\NormalTok{(Xvars, Xtype, }\DataTypeTok{T=}\DecValTok{200}\NormalTok{, }\DataTypeTok{L=}\DecValTok{2}\NormalTok{)}
\end{Highlighting}
\end{Shaded}

\vspace{6pt}\noindent\textbf{ii) Fit the data}

The fitting step is now straight-forward.  The forest is given to the fitting function, together with the data matrix and the responses, either encrypted or unencrypted.  The only additional option is what size of stochastic fraction estimate to use (the $M$ parameter from \S\ref{sec:RFfrac}), passed as argument \texttt{resamp}.

To do this unencrypted, one simply runs:
\begin{Shaded}
\begin{Highlighting}[]
\NormalTok{> fit <-}\StringTok{ }\KeywordTok{CRF.fit}\NormalTok{(forest, X, y, }\DataTypeTok{resamp=}\DecValTok{8}\NormalTok{)}
\end{Highlighting}
\end{Shaded}
If \texttt{X} and \texttt{y} are encrypted, then the algorithm requires the value $1$ encrypted, as well as $M$ both encrypted and in plain text, since the stochastic fraction estimate is $M-\sum_{i=1}^M \eta_i + 1$, where the $\eta_i$ come from \texttt{X}.  
Therefore, the public key must be provided so that the algorithm can produce encrypted versions of these,
\begin{Shaded}
\begin{Highlighting}[]
\NormalTok{> fit <-}\StringTok{ }\KeywordTok{CRF.fit}\NormalTok{(forest, X, y, }\DataTypeTok{resamp=}\DecValTok{8}\NormalTok{, }\DataTypeTok{pk=}\NormalTok{keys\$pk)}
\end{Highlighting}
\end{Shaded}
For the \citet{Fan12} scheme in particular, it is possible to add unencrypted values to encrypted ones knowing only the parameters of the scheme, so this requirement may be removed in a future release of the package.

\vspace{6pt}\noindent\textbf{iii) Prediction}

Imagine now that there are $N'$ test observations for the toy example, similarly stored in an unencrypted or encrypted $N' \times P$ matrix, \texttt{newX}.  Then prediction proceeds using the forest and fit from the previous commands:
\begin{Shaded}
\begin{Highlighting}[]
\NormalTok{> pred <-}\StringTok{ }\KeywordTok{CRF.pred}\NormalTok{(forest, fit, newX)}
\end{Highlighting}
\end{Shaded}
For similar reasons to above, the public key is required when computing with encrypted data,
\begin{Shaded}
\begin{Highlighting}[]
\NormalTok{> pred <-}\StringTok{ }\KeywordTok{CRF.pred}\NormalTok{(forest, fit, newX, }\DataTypeTok{pk=}\NormalTok{keys\$pk)}
\end{Highlighting}
\end{Shaded}
Upon completion, \texttt{pred} will be an $N' \times 2$ matrix with votes for class $j$ on observation $i$ being in the $(i,j)$th entry.

\subsubsection*{Semi-parametric Na\"{i}ve Bayes}

There are two steps in working with the SNB classifier in the package: i) fit a model to the data, i.e., estimate the factors that make up the parameters 
$\{ \hat{\theta}, \hat{\alpha} , \hat{\beta} \}$; ii) predict a new observation.

The user has the choice of working with encrypted or unencrypted data (this is automatically detected), with the provision that both steps must use the same type of data, i.e., either both use encrypted data or both use unecrypted data.

Assume we have an $N \times P$ design matrix \texttt{X} and a length $N$ binary response vector \texttt{y}; and let \texttt{cX} and \texttt{cy} denote the corresponding ciphertexts, encrypted with the package \texttt{HomomorphicEncryption} (see \S\ref{App:HEpkg}).

\vspace{6pt}\noindent\textbf{i) Estimation}

Besides the data there are only two parameters in the estimation step: 
\texttt{paired} controls whether paired or unpaired optimisation is to be used (see \S\ref{sec:stats_glm});
and \texttt{pk} is the public key used to encrypt \texttt{cX} and \texttt{cy} --- it is used to encrypt some constants required in the estimation process and is only needed if encrypted data is used.

Estimating the class counts $\{ \text{counts}(y=0), \text{counts}(y=1) \}$ and the coefficients $\{ a_{j}, b_{j}, d_{j}\}$ and is then straightforward using the fitting function \texttt{SNB.fit}.
Say $N=10$ and $P=2$; then, with unencrypted data we obtain
\begin{Shaded}
\begin{Highlighting}[]
\NormalTok{> snb1.fit <-}\StringTok{ }\KeywordTok{SNB.fit}\NormalTok{(X, y, }\DataTypeTok{paired=}\NormalTok{T)}
\NormalTok{> snb1.fit}
\NormalTok{\$counts.y}
\NormalTok{[1] 4 6}
\NormalTok{\$coeffs}
\NormalTok{  [,1] [,2]}
\NormalTok{a  110  462}
\NormalTok{b   -8 -120}
\NormalTok{d  221  105}
\end{Highlighting}
\end{Shaded}
\noindent
while with encrypted data we obtain
\begin{Shaded}
\begin{Highlighting}[]
\NormalTok{> snb2.fit <-}\StringTok{ }\KeywordTok{SNB.fit}\NormalTok{(cX, cy, }\DataTypeTok{paired=}\NormalTok{T, }\DataTypeTok{pk=}\NormalTok{keys\$pk)}
\NormalTok{> snb2.fit}
\NormalTok{\$counts.y}
\NormalTok{Vector of 2 Fan and Vercauteren cipher texts}
\NormalTok{\$coeffs}
\NormalTok{Matrix of 3 x 2 Fan and Vercauteren cipher texts}
\end{Highlighting}
\end{Shaded}
We can then confirm that encrypted and decrypted results match:
\begin{Shaded}
\begin{Highlighting}[]
\NormalTok{> }\KeywordTok{dec}\NormalTok{(keys\$sk, snb2.fit\$counts.y) ==}\StringTok{ }\NormalTok{snb1.fit\$counts.y}
\NormalTok{[1] TRUE TRUE}
\NormalTok{> }\KeywordTok{dec}\NormalTok{(keys\$sk, snb2.fit\$coeffs)  ==}\StringTok{ }\NormalTok{snb1.fit\$coeffs}
\NormalTok{  [,1] [,2]}
\NormalTok{a TRUE TRUE}
\NormalTok{b TRUE TRUE}
\NormalTok{d TRUE TRUE}
\end{Highlighting}
\end{Shaded}

\vspace{6pt}\noindent\textbf{ii) Prediction}

The function \texttt{SNB.fit} generates an object of class \texttt{SNB}, which can be directly fed into the base R \texttt{predict} function.
Apart from the fitted \texttt{model} and the testing data \texttt{newX}, there are only two parameters:
\texttt{type} controls the type of prediction to the returned (raw coefficients or conditional class probabilities);
\texttt{sk} is the secret key --- required if predictions are to be decrypted.

Continuing the example above, we can obtain the probabilities of class $y=1$,
\begin{Shaded}
\begin{Highlighting}[]
\NormalTok{> snb2.predProb <-}\StringTok{ }\KeywordTok{predict}\NormalTok{(}\DataTypeTok{model=}\NormalTok{snb.fit2, }\DataTypeTok{newX=}\NormalTok{cX, }\DataTypeTok{type=}\StringTok{"prob"}\NormalTok{, }\DataTypeTok{sk=}\NormalTok{keys\$sk)}
\NormalTok{> snb2.predProb}
\NormalTok{[1] 0.2212818 0.1973391 0.7221404 0.8975560 0.7293452 0.4353264 0.4621838}
\NormalTok{[8] 0.4442447 0.7364313 0.8907048}
\end{Highlighting}
\end{Shaded}
\noindent
or obtain the class counts and coefficients $\{ e_{ij}, d_{j} \}$ (see \S\ref{sec:stats_glm}) for all testing observations,
\begin{Shaded}
\begin{Highlighting}[]
\NormalTok{> snb2.predRaw <-}\StringTok{ }\KeywordTok{predict}\NormalTok{(}\DataTypeTok{model=}\NormalTok{snb.fit2, }\DataTypeTok{newX=}\NormalTok{cX, }\DataTypeTok{type=}\StringTok{"raw"}\NormalTok{, }\DataTypeTok{sk=}\NormalTok{keys\$sk)}
\NormalTok{> snb2.predRaw}
\NormalTok{\$counts.y}
\NormalTok{[1] 4 6}
\NormalTok{\$coeffs.e}
\NormalTok{      [,1] [,2]}
\NormalTok{ [1,]  102 -138}
\NormalTok{ [2,]   70 -138}
\NormalTok{ [3,]   86  102}
\NormalTok{ [4,]  102  222}
\NormalTok{ [5,]   94  102}
\NormalTok{ [6,]   70  -18}
\NormalTok{ [7,]   94  -18}
\NormalTok{ [8,]   78  -18}
\NormalTok{ [9,]  102  102}
\NormalTok{[10,]   86  222}
\NormalTok{\$coeffs.d}
\NormalTok{[1] 221 105}
\end{Highlighting}
\end{Shaded}

\newpage

\section{Details of data sets}
\label{sec:datasets}

\begin{table}[!h]
\captionsetup{width=\textwidth}
\caption{Dimensions of the data sets used (shown in the same order as in Figure~\ref{fig:performance}). All data sets were extracted from UCI Machine Learning Repository \citep{Lichman13}.\hfill{ }}\vspace{-0.5cm} 
\label{tab:datasets}
\begin{small}
\begin{center}
\begin{tabular*}{\textwidth}{l @{\extracolsep{\fill}} ccccc}
\hline\hline 

&  \textbf{\#{obs}}$_{1}$
&  \textbf{\#{obs}}$_{2}$
&  \textbf{\#{vars}}$_{1}$
&  \textbf{\#{vars}}$_{2}$
&  \textbf{\#{vars}}$_{3}$ \\
\hline

\texttt{acute inflammation}
& 120 & 120 & 6 & 11 & 15 \\
\texttt{acute nephritis}
& 120 & 120 & 6 & 11 & 15 \\
\texttt{adult income}
& 48842 & 45222 & 14 & 105 & 129 \\
\texttt{bank marketing}
& 41188 & 30488 & 20 & 57 & 97 \\
\texttt{blood transfusion}
& 748 & 748 & 3 & 3 & 15 \\
\texttt{breast cancer wisc diag}
& 569 & 569 & 30 & 30 & 150 \\
\texttt{breast cancer wisc orig}
& 699 & 683 & 9 & 9 & 45 \\
\texttt{breast cancer wisc prog}
& 198 & 194 & 32 & 32 & 160 \\
\texttt{chess krvkp}
& 3196 & 3196 & 36 & 73 & 73 \\
\texttt{haberman survival}
& 306 & 306 & 3 & 14 & 22 \\
\texttt{heart disease cleveland}
& 303 & 297 & 13 & 28 & 48 \\
\texttt{ionosphere}
& 351 & 351 & 32 & 32 & 160 \\
\texttt{magic telescope}
& 19020 & 19020 & 10 & 10 & 50 \\
\texttt{mammographic masses}
& 961 & 830 & 4 & 14 & 18 \\
\texttt{monks3}
& 554 & 554 & 6 & 17 & 17 \\
\texttt{musk1}
& 476 & 476 & 166 & 166 & 830 \\
\texttt{musk2}
& 6598 & 6598 & 166 & 166 & 830 \\
\texttt{ozone 1hr}
& 2536 & 1848 & 72 & 72 & 360 \\
\texttt{spambase}
& 4601 & 4601 & 57 & 57 & 285 \\

\hline
\end{tabular*}
\end{center}
\end{small}
\vspace{-10pt}
{\fontsize{8pt}{1em}\selectfont 
\#{obs}$_{1}$: number of observations in the original dataset;
\#{obs}$_{2}$: number of observations after removing missing data;
\#{vars}$_{1}$: number of predictors in the original dataset;
\#{vars}$_{2}$: number of predictors after transforming factors into sets of binaries (which increases the number of predictors) and continuous predictors into quintile-discretised predictors (which does not increase the number of predictors);
\#{vars}$_{3}$: number of predictors after also transforming quintile predictors into sets of binaries.
}
\end{table}

\newpage
\section{Completely random forest parameter performance}
\label{sec:CRFperformance}

\begin{figure}[!ht]
\captionsetup[subfigure]{belowskip=-1cm}
\centering
\includegraphics[width=1.2\textwidth,angle=-90,origin=c]{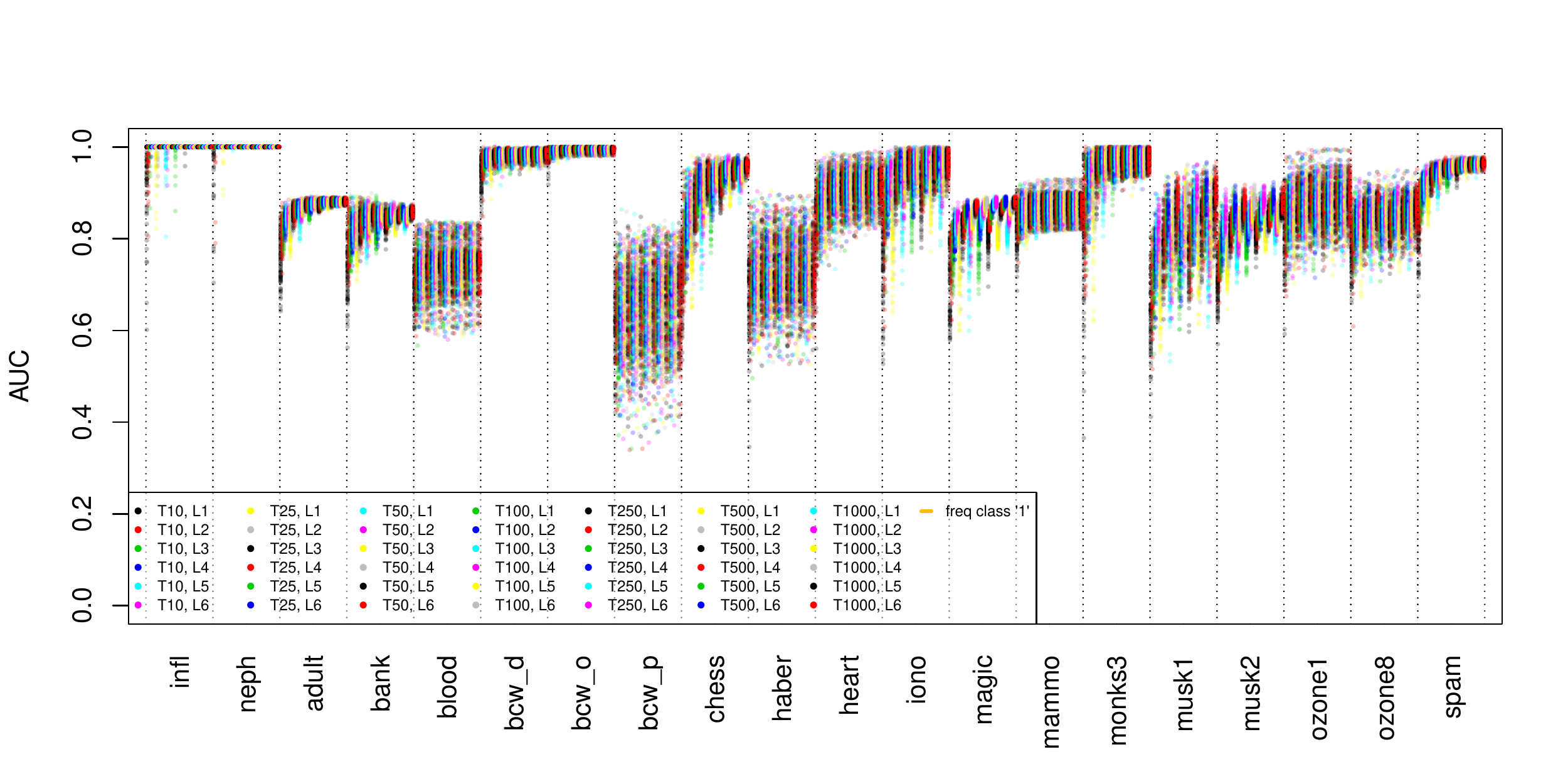}
\vspace{-0.5cm}
\caption{Performance of various methods. For each model and dataset, the AUC for 100 stratified randomisations of the training and testing sets; the horizontal lines represent the frequency of class $y=1$.  $M=8$ throughout.}
\label{fig:CRFperformance}
\end{figure}

\end{document}